\documentclass[journal]{IEEEtran}

\usepackage{amsmath,amsfonts}
\usepackage{algorithm}
\usepackage{array}
\usepackage[caption=false,font=normalsize,labelfont=sf,textfont=sf]{subfig}
\usepackage{textcomp}
\usepackage{stfloats}
\usepackage{url}
\usepackage{verbatim}
\usepackage{graphicx}
\usepackage{cite}
\usepackage{xcolor}
\usepackage{hyperref}
\hypersetup{colorlinks=true, linkcolor=blue, linktocpage}
\usepackage[noEnd]{algpseudocodex}
\usepackage{indentfirst}
\usepackage{esvect}
\usepackage{orcidlink}
\usepackage{booktabs}       
\usepackage{tabularx}                     
\usepackage{multirow, makecell}
\newcolumntype{K}{>{\arraybackslash}p{0.0352\textwidth}}
\definecolor{gold}{HTML}{ffbf00}
\definecolor{silver}{HTML}{b4b4b4}
\definecolor{bronze}{HTML}{a98565}
\definecolor{lightgreen}{HTML}{90ee90}
\definecolor{lightred}{HTML}{f08080}
\usepackage{makecell}
\usepackage[export]{adjustbox}
\usepackage{txfonts}
\usepackage{siunitx}

\begin{document}
\title{ A new Image Similarity Metric for a Perceptual and Transparent Geometric and Chromatic Assessment}

\author{\IEEEauthorblockN{Antonio Di Marino\IEEEauthorrefmark{1} \orcidlink{0000-0001-5748-2838}}\IEEEcompsocitemizethanks{\IEEEcompsocthanksitem\IEEEauthorrefmark{1}These authors share the first authorship.}, Vincenzo Bevilacqua\IEEEauthorrefmark{1} \orcidlink{0000-0002-7139-3222}, Emanuel Di Nardo \orcidlink{0000-0002-6589-9323}, Angelo Ciaramella \orcidlink{0000-0001-5592-7995}, \\Ivanoe De Falco \orcidlink{0000-0001-6127-1195}, Giovanna Sannino \orcidlink{000-0001-7856-8761}
\thanks{Manuscript received April 19, 2021; revised August 16, 2021.}
\thanks{\textit{(Corresponding author: Vincenzo Bevilacqua.)}
}
\thanks{Antonio Di Marino and Vincenzo Bevilacqua are with the Institute for High-Performance Computing and Networking (ICAR) - National Research Council (CNR), Naples, Italy, and also with the University of Naples Federico II, Department of Electrical Engineering and Information Technology, via Claudio 21, Naples, 80125, Italy (email: antonio.dimarino@icar.cnr.it; \,vincenzo.bevilacqua@icar.cnr.it). }

\thanks{Ivanoe De Falco and Giovanna Sannino are with Institute for High-Performance Computing and Networking (ICAR) - National Research Council (CNR), Naples, Italy (email: ivanoe.defalco@icar.cnr.it; \,giovanna.sannino@icar.cnr.it).}
\thanks{Emanuel Di Nardo and Angelo Ciaramella are with University of Naples Parthenope, Naples, Italy (email: emanuel.dinardo@uniparthenope.it; \;angelo.ciaramella@uniparthenope.it).}

\thanks{The official implementation of our metric is publicly available at {https://github.com/antdimarino/EDOKS}.}}

\markboth{Journal of \LaTeX\ Class Files,~Vol.~14, No.~8, August~2021}%
{Shell \MakeLowercase{\textit{et al.}}: A Sample Article Using IEEEtran.cls for IEEE Journals}
\IEEEpubid{
    \begin{minipage}{\textwidth}\centering
        \vspace{12pt}
        This work has been submitted to the IEEE for possible publication. \\
        Copyright may be transferred without notice, after which this version may no longer be accessible.
    \end{minipage}
}

\maketitle

\begin{abstract}
In the literature, several studies have shown that state-of-the-art image similarity metrics are not perceptual metrics; moreover, they have difficulty evaluating images, especially when texture distortion is also present. In this work, we propose a new perceptual metric composed of two terms. The first term evaluates the dissimilarity between the textures of two images using Earth Mover's Distance. The second term evaluates the chromatic dissimilarity between two images in the Oklab perceptual color space. We evaluated the performance of our metric on a non-traditional dataset, called Berkeley-Adobe Perceptual Patch Similarity, which contains a wide range of complex distortions in shapes and colors. We have shown that our metric outperforms the state of the art, especially when images contain shape distortions,  confirming also its greater perceptiveness. Furthermore, although deep black-box metrics could be very accurate, they only provide similarity scores between two images, without explaining their main differences and similarities. Our metric, on the other hand, provides visual explanations to support the calculated score, making the similarity assessment transparent and justified.
\end{abstract}


\section{Introduction}

\IEEEPARstart{I}{n} image processing and analysis, the use of a similarity metric is essential when the goal is to quantify the similarity or dissimilarity between two images. This is useful in many tasks, such as evaluating a lossy compression algorithm or evaluating image generative models, where the goal is to evaluate the quality of an artificially generated image against the target image by providing a score.

The present study examines Image Quality Assessment (IQA) metrics from a perceptual perspective, understood as the ability of the metrics to assess image similarity based on the characteristics and cognitive mechanisms of the Human Visual System (HVS) \cite{watson_digital_1993}. 

\begin{figure}[!t]
\includegraphics[width=\linewidth]{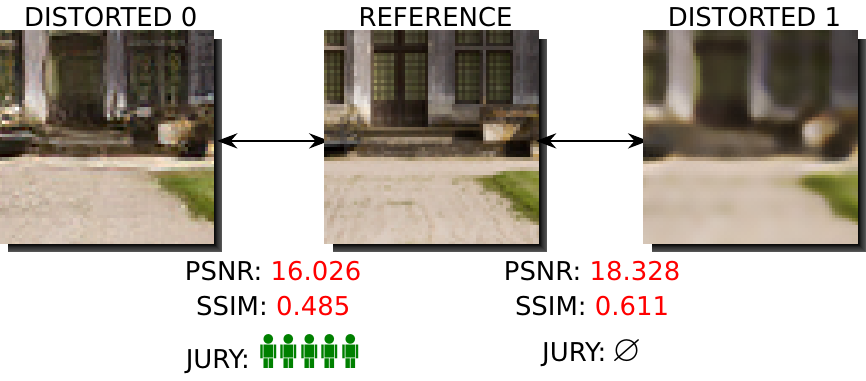}
\caption{Comparison of human perceptual similarity and nonperceptual metrics. On a sample of the 2AFC dataset \cite{zhang_unreasonable_2018}, the similarity metrics PSNR and SSIM \cite{wang_mean_2009} assigned the highest similarity score between the reference image and the distorted one on the right, suggesting that, for both metrics, the reference image is more similar to the distorted image on the right than the distorted image on the left. However, a panel of five human judges unanimously agreed that based on their visual perception, the reference image is more similar to the left distorted image.}
\label{fig:perceptual}
\end{figure}

In the current state-of-the-art (SOTA), the most widely used similarity metrics are the Peak Signal-to-Noise Ratio (PSNR) and the Structural Similarity Index Measure (SSIM) \cite{wang_image_2004}. 
However, as shown in Figure \ref{fig:perceptual}, they are not perceptual in nature. This is due to the fact that they do not align with human perception when it comes to assessing the similarity between two sets of images.

This has also been documented in several works, such as the work of Ebrahimi et al. \cite{ebrahimi_jpeg_2004}, which shows that PSNR is not a good perceptual metric. In the work of Wang et al. \cite{zhou_wang_no-reference_2002}, the authors show that PSNR is a poor indicator of subjective quality in the evaluation of JPEG compressed images. As an alternative, the authors propose a No-Reference quality measurement algorithm. The work of Wang et al. \cite{wang_mean_2009}, is the most comprehensive work in highlighting the difficulties of Mean Squared Error (MSE) as a measure of signal fidelity. The authors demonstrated the limitations of MSE by comparing an image with its distorted versions, measuring the distance between each pair of images, and showing that this metric does not accurately reflect signal fidelity. Since PSNR uses MSE in its equation, PSNR inherits these same problems.

\IEEEpubidadjcol

Since PSNR has been shown not to be a perceptual metric, the SSIM metric was subsequently proposed; there are works in the literature showing that there are correlations between these two metrics, and furthermore, demonstrating that SSIM can also produce unexpected scores. Horé and Ziou \cite{hore_image_2010} showed that there is a direct mathematical relationship between PSNR and SSIM, demonstrating that PSNR can be derived as a function of SSIM. In addition to proving the mathematical connection, the authors performed empirical experiments to further demonstrate the similarity between PSNR and SSIM.

In recent years, SSIM has become the most widely used metric in a wide range of contexts. However, in the work of Nilsson and Akenine-Möller \cite{nilsson_understanding_2020}, the authors analyze the mathematical factors of SSIM and demonstrate that SSIM can produce unexpected, often undefined and non-intuitive results leading the user to erroneous conclusions; they affirm that this metric is being used for more than what it was created for. 

We have observed that despite the innovative high-level IQA metrics being closer to HVS \cite{zhang_unreasonable_2018}, both PSNR and SSIM are now used as de facto standards 
\cite{ma2020structure,ma2021structure,lim_enhanced_2017,zhang_image_2018,dai_second-order_2019,niu_single_2020,mei_image_2021}, although the various works mentioned above have shown how in different contexts they are not the most appropriate metrics. 

On the other hand, high-level deep learning-based metrics, such as Learned Perceptual Image Patch Similarity (LPIPS) \cite{zhang_unreasonable_2018}, have shown high correlations with human judgment; however, their inherent black-box nature significantly limits their transparency and interpretability, obscuring the specific visual components that determine similarity or difference. 

In response to this limitation, this work proposes the \textit{Earth mover's Distance and Oklab Similarity} (EDOKS) metric, which is based on the integration of low-level handcrafted features. This analytical approach is specifically designed to transparently and distinctly quantify the impact of both geometric (shape) and chromatic (color) distortions. 
Given the analytical and white-box nature of EDOKS, and considering that deep learning-based metrics achieve extremely high accuracy at the expense of transparency, our goal in this paper is to demonstrate that, compared to other low-level IQA metrics, EDOKS is more consistent with human perceptual evaluation.

We propose two terms of perceptual dissimilarity for evaluating differences between two images, one related to textures and the other to colors. Furthermore, we utilize these terms to attempt to define a unique index of perceptual similarity. 
The first term related to the texture is calculated using the Earth Mover's Distance (EMD) \cite{rubner_earth_2000}, while the second term calculates the distances between colors in the Oklab color space \cite{ottosson_perceptual_2020}. Finally, we propose the combination of these two terms creating the EDOKS index. 

The vast majority of low-level SOTA metrics have been evaluated and optimized on datasets containing chromatic, noise, and a few compression distortions. We, instead, use the Berkeley-Adobe Perceptual Patch Similarity (BAPPS) Dataset \cite{zhang_unreasonable_2018} because it contains more distortions, not only the classic ones, but also those generated by deep models, as well as distortions in shapes, textures, and other aspects.

Performing several experiments, we demonstrate that EDOKS behaves consistently with the perceptual similarity of color and textures between images, a behavior not always guaranteed by SOTA similarity metrics.

Furthermore, given the lack of transparency in all SOTA IQA metrics, especially deep ones based on black-box architectures, we show that our metric is, instead, easily explainable. It provides maps of interest that highlight areas with significant differences between two images, ensuring transparency in its use.

\section{Related Works}\label{sec:related_works}

Metrics for evaluating the quality of an image can be classified into three different categories \cite{shahid_no-reference_2014}:
\begin{itemize}
    \item \textit{Full-Rreference} (FR) methods evaluate the quality of an image by comparing it with the corresponding original target, so this class of metrics can only be used if the original signal is available.
    \item \textit{Reduced-Reference} (RR) methods are used when one does not have direct access to the original target images, but has a reduced and limited feature set.
    \item \textit{No-Reference} (NR) methods are used when the target image is not available; these are usually approximate methods and less accurate than the other classes of methods.
\end{itemize}

In this paper, we propose a metric that belongs to the class of FR methods; in the current SOTA, the most widely used FR similarity metrics are the PSNR and the SSIM. 

The PSNR was originally created to evaluate the quality of images in compression of lossy type, thus with loss of information \cite{rabbani_overview_2002}. 
PSNR calculates the ratio of the maximum pixel value to the MSE between two images, and expresses the result as a logarithmic quantity on the decibel (dB) scale. PSNR is based on local differences between pixels and is unable to evaluate regional information such as texture or shape.

The SSIM is proposed by Wang et al. \cite{wang_image_2004} as an alternative perceptual metric to PSNR, and is defined as a perceptual metric.
In addition to evaluating luminance and contrast between two images, it also compares the structural information of the objects in the images because pixels have strong spatial dependencies.

In the basic version, SSIM is applied to only one channel, typically luminance. However, it can also be applied to RGB images or other color spaces. In this case, SSIM is calculated independently on each channel and then combined into a single score using a weighted sum \cite{wang_video_2004}. This method does not consider the more robust chromatic information.

Since the metrics just mentioned have been the standard for image quality assessment for years, several papers have proposed enhancements of PSNR and SSIM to try to achieve more accurate similarity assessments. For the PSNR, the PSNR Human Visual System Masking model (PSNR-HVS-M) \cite{ponomarenko2007between} was proposed, which consists of the calculation of the MSE between block Discrete Cosine Transform (DCT) coefficients; even in this version of PSNR, structural information such as shapes or textures is not considered. Instead of SSIM, Multi-Scale SSIM (MS-SSIM) \cite{wang_multiscale_2003} was introduced as an enhancement, since SSIM strongly depends on the scale at which it is applied. It is proposed that SSIM be evaluated at different scales. The use of different scales allows image details to be evaluated at different resolutions. Another enhancement is Information Content Weighting SSIM (IW-SSIM)  \cite{zhou_wang_information_2011}, where the authors combine information content weighting with MS-SSIM. But even in these two latter cases, the metric is based on the evaluation of luminance between two images.

Other FR methods have been proposed in the literature. Firstly, we can recall here the feature similarity index (FSIM) \cite{lin_zhang_fsim:_2011}, that employs two features, the phase congruency and the gradient magnitude, to compute the local similarity map.

The authors claimed that the phase congruency and the gradient magnitude play complementary roles in characterizing the local image quality. At the quality score pooling stage of FSIM, phase congruency map is utilized again as a weighting function since it can roughly reflect how perceptually important a local patch is to the HVS, but it is not a metric designed to be robust to distortions in shapes or textures. Spectral Residual-based similarity (SR-SIM) \cite{zhang_sr-sim:_2012} is a metric that stands as a computationally faster alternative to IW-SSIM and FSIM. It consists of using spectral residual visual saliency both to extract feature maps on local image qualities and as a weight of the final score, but ignores color features and geometric shapes.

Visual Information Fidelity (VIF) metric \cite{sheikh_image_2006} models the image as a source of information passed through two channels, one representing the distortions introduced by the acquisition or compression system, and one modeling the HVS. VIF then calculates the mutual information between the original and distorted images in a wavelet domain. It does not distinguish well where the distortion occurs because it operates globally in the wavelet domain, and it was tested on simple datasets and fixed models, ignoring realistic distortion scenarios or new types of artifacts \cite{cao_image_2024}.

Gradient Magnitude Similarity Deviation (GMSD) \cite{xue_gradient_2014} is a metric that makes use of gradient-based similarity using only the luminance component, that is, it uses image gradients and a new pooling strategy to analyze similarity between images. Therefore, it is unable to effectively capture multiscale perceptual gradients or chromatic distortions. Multiscale and extended color versions of GMSD (MS-GMSD) \cite{zhang_gradient_2017} were subsequently proposed precisely to overcome these limitations. However, both metrics are based exclusively on gradient maps and do not feature any explicit modeling of shape or structural coherence, thus remaining insensitive to geometric or spatial misalignments.

Visual Saliency-based Index (VSI) \cite{zhang_vsi:_2014} is a metric that uses saliency maps not only as a weight function for score pooling, but also as feature maps to characterize local image quality. VSI assumes that the chosen saliency map is suitable and stable for all types of distortions. However, the perception of saliency may change in the presence of certain artifacts or in complex visual contexts. This means that metrics such as VSI which assume standard conditions may have gaps in the following cases: extreme ambient light conditions, poorly calibrated displays, viewing in highly reflective environments, etc. \cite{chubarau_perceptual_2020}. 

In the similarity of DCT subbands (DSS) \cite{balanov_image_2015}, the important features of human perception are measured by the variation of structural information in the subbands of the DCT domain. DSS' authors place greater weight on low subbands, so if the distortion is localized in a high band or manifests itself more as a localized artifact, the metric may “lose” sensitivity. Furthermore, DSS calculates contributions per subband and does not strongly incorporate spatial localization or the visual importance of regions.

Mean Deviation Similarity Index (MDSI) \cite{ziaei_nafchi_mean_2016} is a metric that consists of three terms: gradient similarity, chromaticity similarity and deviation pooling. MDSI uses local gradients on a single spatial scale and may fail to detect geometric distortions that alter the overall structure. Therefore, although it is effective for blurring, noise, and local contrast degradation, its sensitivity to global structural changes is limited. 

In \cite{reisenhofer_haar_2018}, the authors proposed Haar wavelet-based Perceptual Similarity Index (HaarPSI) a metric that uses the coefficients resulting from the Haar wavelet decomposition to evaluate the similarity between two images. This is a good similarity metric, similar to our approach in that it uses filters to extract information from the image. However, one limitation is its sensitivity to preprocessing, which causes the metric to behave unreliably \cite{siniukov_applicability_2023}.

For all the metrics in which coefficient optimization is required, the authors execute this process on datasets that do not adequately represent geometric distortions and artifacts generated by deep neural networks. Consequently, these methods are ineffective for contemporary applications, such as evaluating the output of deep models. 

Furthermore, they were not developed with the intention of being transparent and explainable. They do not show which visual elements had the greatest impact on the computation of the similarity score between two images.

In generative AI models, where the generated artificial image must be compared with the target image, other FR similarity metrics have also been proposed. However, these metrics utilize black-box models to extract image features and evaluate the similarity between images. This makes it difficult to trace the elements and their positions in the image that led to the discrimination of the similarity score. 

The primary similarity metric that leverages a deep architecture is the LPIPS \cite{zhang_unreasonable_2018}, that measures the distance between two images in the feature space of deep neural networks pre-trained on visual recognition tasks. The activations of features from multiple layers of the network are compared and combined in a linear fashion using weights that have been learned. The purpose of this process is to approximate human perceptual judgments.

Unlike LPIPS, Perceptual image-error Assessment through Pairwise Preference (PieAPP) \cite{prashnani_pieapp:_2018} is trained on paired human judgments: the model learns to predict which of the two distorted images is perceptually closer to the target image. In this way, the neural network estimates a “perceptual error” consistent with human visual perception.

Another deep FR perceptual metric is Deep Image Structure and Texture Similarity (DISTS) \cite{ding_image_2020}, which combines structure and texture using deep network features. Compared to LPIPS, it explicitly separates contributions from structures and textures, improving consistency with human perception and stability across different distortions.

Finally, TOPIQ \cite{chen_topiq:_2024} uses a Transformer-based top-down approach that starts with semantic understanding of the image and ends with the evaluation of local distortions.

An alternative approach to the similarity metrics already mentioned is to use a human jury to evaluate the generated images, in which a similarity score must be assigned between two images without knowing which image is generated; this approach is known as Mean Opinion Score (MOS) metrics \cite{menon_pulse:_2020, kim_progressive_2019, ledig_photo-realistic_2017}.

\begin{figure*}[!tb]
\includegraphics[width=\linewidth]{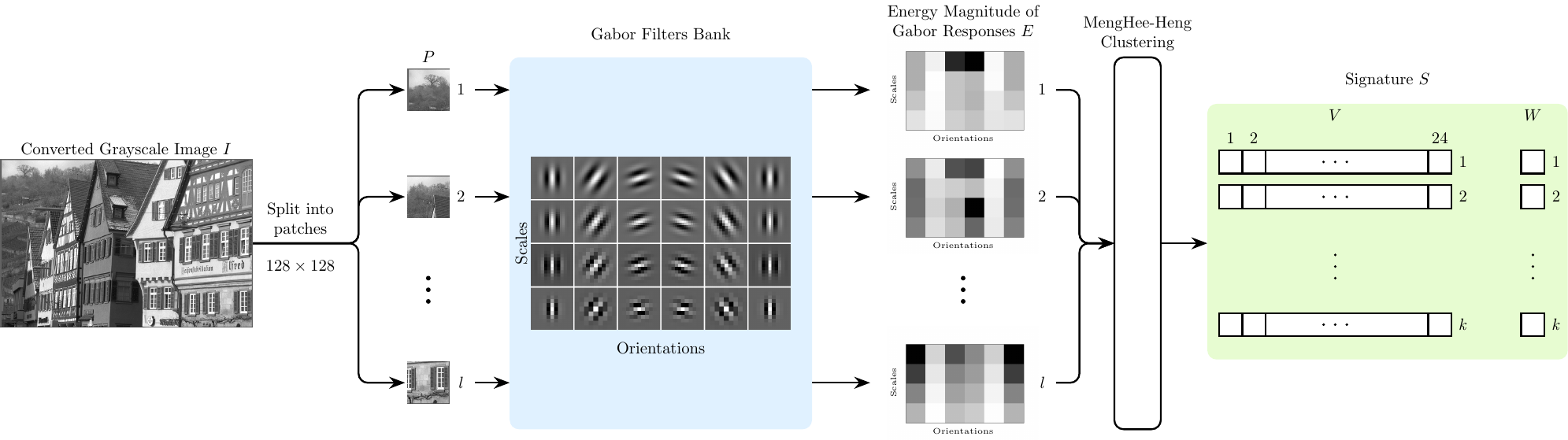}
\caption{The following example illustrates the process of creating a signature from an image. The image is converted to grayscale and split into patches. Subsequently, Gabor's filter dictionary is applied to each patch. The energy magnitudes are calculated for each Gabor response, resulting in a 4x6 matrix for each patch. To reduce the number of patches, the MengHee-Heng clustering algorithm is applied to the matrices, yielding k 24-dimensional centroids V and the corresponding frequencies W. V and W are then output to the S dictionary.}
\label{signature}
\end{figure*}  

\section{Perceptual Terms}
.
From the works in the literature, discussed in the previous section, the lack of terms that can express a perceptual and transparent comparison between the textures and colors of two images stands out. Our goal is to have a perceptual similarity metric that is capable of justifying its score while ensuring transparency. In this section, we analyze the term based on the EMD to compare the dissimilarity between textures in two images; the term exploiting the Oklab color space to compare the perceptual dissimilarity between colors in two images; and EDOKS index as a combination of the two terms. 

\subsection{Texture Dissimilarity Term}

To compare the textures of two images, we relied on the work of Rubner et al. \cite{rubner_earth_2000}, where the authors propose an image retrieval method based on the use of EMD, that is a distance between two distributions used by Hitchcock to solve the Monge-Kantorovich transport problem \cite{hitchcock_distribution_1941}. Histograms are typically used to represent the distribution of an image. In \cite{rubner_earth_2000}, it is shown that histograms are fixed-size structures and therefore do not have balance between expressiveness and efficiency. In fact, the authors propose the use of variable-size image signatures, obtained by clustering the responses achieved using a Gabor filter dictionary applied to image patches. 

In our work, we exploited some steps of \cite{rubner_earth_2000} that allowed us to obtain consistent results for texture comparisons. In the Algorithm \ref{alg:signature}, we show pseudocode of how we obtain signatures $S$ by applying Gabor's filter dictionary to an image. Given an input colored image of size $M \times N \times 3$, we convert it to grayscale $I$ with size $M \times N$, because to evaluate shape dissimilarity it is sufficient to use only the intensity of pixels; moreover, we provide a dedicated term for color evaluation later. We divide the image into $l$ non-overlapping patches of size $p \times p$ where $p << min(M,N)$, and apply Gabor's filter dictionary \cite{gabor_theory_1946} to each patch.

\begin{algorithm}[!b]
\caption{Texture signature extraction}
\label{alg:signature}
\begin{algorithmic}[1]
\Function{signature}{$Image$ $I$}
    \State $s \leftarrow \textrm{[$0.1,0.2,0.3,0.4$]}$  \Comment{list of scales}
    
    \State $o \leftarrow \textrm{[$0^{\circ}, 30^{\circ}, 60^{\circ}, 90^{\circ}, 120^{\circ}, 150^{\circ}$]}$ \Comment{list of orientations}
    \State $P \leftarrow$ \textrm{$patches(I)$} \Comment{non-overlapped patches list}\smallskip
    \State $l \leftarrow length(P)$
    \State $E \leftarrow \textrm{empty\_matrices($l,4,6$)}$  \Comment{energies list}\smallskip
    \For {$z \leftarrow$ 1 \textbf{to} $l$}
        \State $a \leftarrow 0$     \Comment{accumulator}
        \For { $i \leftarrow$ 1 \textbf{to} $length(s)$ }
            \For { $j \leftarrow$ 1 \textbf{to} $length(o)$ }\smallskip
                \State $F_{real}, F_{imag} \leftarrow \textrm{$gabor$($P_z$, $s_i$, $o_j$)}$\smallskip
                \State $|F| \leftarrow \sqrt{F_{real}^2 + F_{imag}^2}$\smallskip
                \State $E_{z,i,j} \leftarrow \textrm{$ sum(|F|^2)$}$
                \State $a \leftarrow a + E_{z,i,j} $
            \EndFor
        \EndFor
        \State $E_{z} \leftarrow  E_{z} / a $             \Comment{Gabor energy}
    \EndFor

    \State $V, W \leftarrow MengHeeHeng(E)$ \Comment{Clustering}

    \State $S \leftarrow  (V, W) $ \Comment{tuple of values and weights}

\State \Return \textrm{$S$} \Comment{return signature of image I}
\EndFunction

\end{algorithmic}
\end{algorithm}

A single Gabor filter allows us to extract the texture features of a patch with respect to a precise scale and orientation. As a result, a dictionary of Gabor filters allows the extraction of texture features at different scales and orientations to try to extract the main patterns of the patch \cite{bovik_multichannel_1990, farrokhnia_multi-channel_1991, manjunath_texture_1996}. In our setup, we define a dictionary of Gabor filters with the following four scales $s = \{0.1,0.2,0.3,0.4\}$ and the following six orientations $o = \{0^{\circ}, 30^{\circ}, 60^{\circ}, 90^{\circ}, 120^{\circ}, 150^{\circ}\}$, which then allows us to obtain a collection of twenty-four total filters for each patch. Each Gabor filter, with a scale $s_i$ and orientation $o_j$, applied to the $z$-th patch produces two responses $F_{real}$ and $F_{imag}$, which are produced by the real and imaginary components of the Gabor kernel convolved with the $z$-th patch, respectively. From the real and imaginary components we calculate the magnitude $|F|$ with the same size of the $z$-th patch $p \times p$. Afterwards, we calculate the energy of the magnitude $|F|$ as in \cite{rubner_perceptual_2001}, corresponding to scale $i$ and orientation $j$, to be assigned to $E_{z,i,j} \in \mathbb{R}$. We get a list of $l$ four-by-six matrices $E$, normalized so that $\sum_{i,j} E_{z,i,j} = 1$ for each $z$-th patch. Each element $E_{z,i,j}$ corresponds to a precise texture response of patch $P_z$ at a scale $s_i$ and orientation $o_j$. 

Figure \ref{signature} illustrates an example of signature creation, where in the list of four-by-six matrices $E$, a dark pixel signifies a robust response to a specific scale and orientation. To use EMD, the distributions must be weighted, so as a final step in creating the signatures we apply a \textit{Meng-Hee Heng clustering} algorithm, which takes as input the $l$ energies and returns as output two lists of $k$ centroids $V = \{v_1, ..., v_k\} \; | \; v_i \in \mathbb{R}^{24}$ and $k$ weights $W = \{w_1, ..., w_k\} \; | \; w_i \in \mathbb{R}$, where $k \leq l$. Each centroid $v_i$ represents a texture patch that appears frequently in the image $I$, $v_i$ is a four-by-six matrix or can be viewed as a single 24-dimensional vector. Each weight $w_i$ represents the frequency percentage of $v_i$ that appears in image $I$. Unlike \cite{rubner_perceptual_2001}, where they use K-means as the clustering algorithm, we preferred the Meng-Hee Heng variant, which automatically finds the suitable number of $k$ clusters. At the end, both the list of 24-dimensional centroids $V$ and the list of corresponding weights $W$ make up the $S$ signature of a single image $I$.  

Let $S_X = (\{v_{x_1},...,v_{x_n}\}, \{w_{x_i},...,w_{x_n}\})$ be the signature obtained from image $X$ with $n$ clusters, and $S_Y = (\{v_{y_1},...,v_{y_m}\}, \{w_{y_i},...,w_{y_m}\})$ the signature obtained from image $Y$ with $m$ clusters, the next step is to calculate the EMD between the two signatures. Commonly in EMD, the two distributions can be seen as two land masses, where the first distribution is uniformly distributed in space and the second distribution is a set of holes in this same space. The goal of EMD is to measure the minimum amount of work to fill the holes in the second distribution using the ground of the first distribution. 

Formally, the EMD is a distance between two weighted distributions $S_X$ and $S_Y$, with the following equation:

\begin{equation}
    EMD(S_X,S_Y) = \frac{\sum_{i=1}^n \sum_{j=1}^m d(v_{x_i}, v_{y_j})f_{ij}}{\sum_{i=1}^n \sum_{j=1}^m f_{ij}}
\end{equation}
where $d(v_{x_i}, v_{y_j})$ is the ground distance between the two centroids of different signatures, by default is the L1 distance; and $f_{ij}$ is an element of flow matrix $F$ computed as in \cite{rubner_earth_2000}.

The score obtained from the EMD equation is a dissimilarity value between the texture signatures of the two images $X$ and $Y$. If the images are equal, the score obtained is $0$, while the more dissimilar the two images are, the more the score tends to increase.

\subsection{Color Perceptual Dissimilarity Term} 

Considering that the EMD term only compares texture signatures between two grayscale images, we believe that a good comparison between two images should be made by relating colors as well. For the color dissimilarity term, our goal is to make this index faithful to how humans perceive colors differences. Since the RGB representation is not uniformly perceptual, over the years, various color spaces have been proposed in image processing that mimic the human perceptual ability to recognize colors and their distributions. 

Our choice as a color space to use falls on Oklab \cite{ottosson_perceptual_2020}, which allows a perceptual color representation in a simple way. The other evaluated perceptual color spaces have the following problems: CIE XYZ \cite{smith_c.i.e._1931} is perceptually non-uniform; CIE Lab \cite{standard2007colorimetry} is approximately uniform, but uniformity and linearity of hue present problems in representing blue \cite{hung_determination_1995, zhao_hue_2020}; LAB2000 \cite{lissner_how_2010} is complex because it uses lookup tables to create a non-euclidean hyperspace by not solving non-uniformity problems in blue regions; IPT \cite{ebner_derivation_1998} is specifically designed to be a linear space, sacrificing uniformity, in representing hues; CAM16-USC \cite{schlomer_algorithmic_2018} is a very complex, numerically unstable, and not always invertible color space.

Oklab, on the other hand, is designed to ensure uniformity and perceptual linearity in hue, so that it does not have problems such as misrepresentation of the blue region. In addition, among its many features, Oklab guarantees that the coordinates are perceptually orthogonal, the space is numerically simple, easily invertible, scale-independent, and it is the color space that best approximates Munsell space\cite{munsell_pigment_1912}.

\begin{figure}[!tb]
\includegraphics[width=\linewidth]{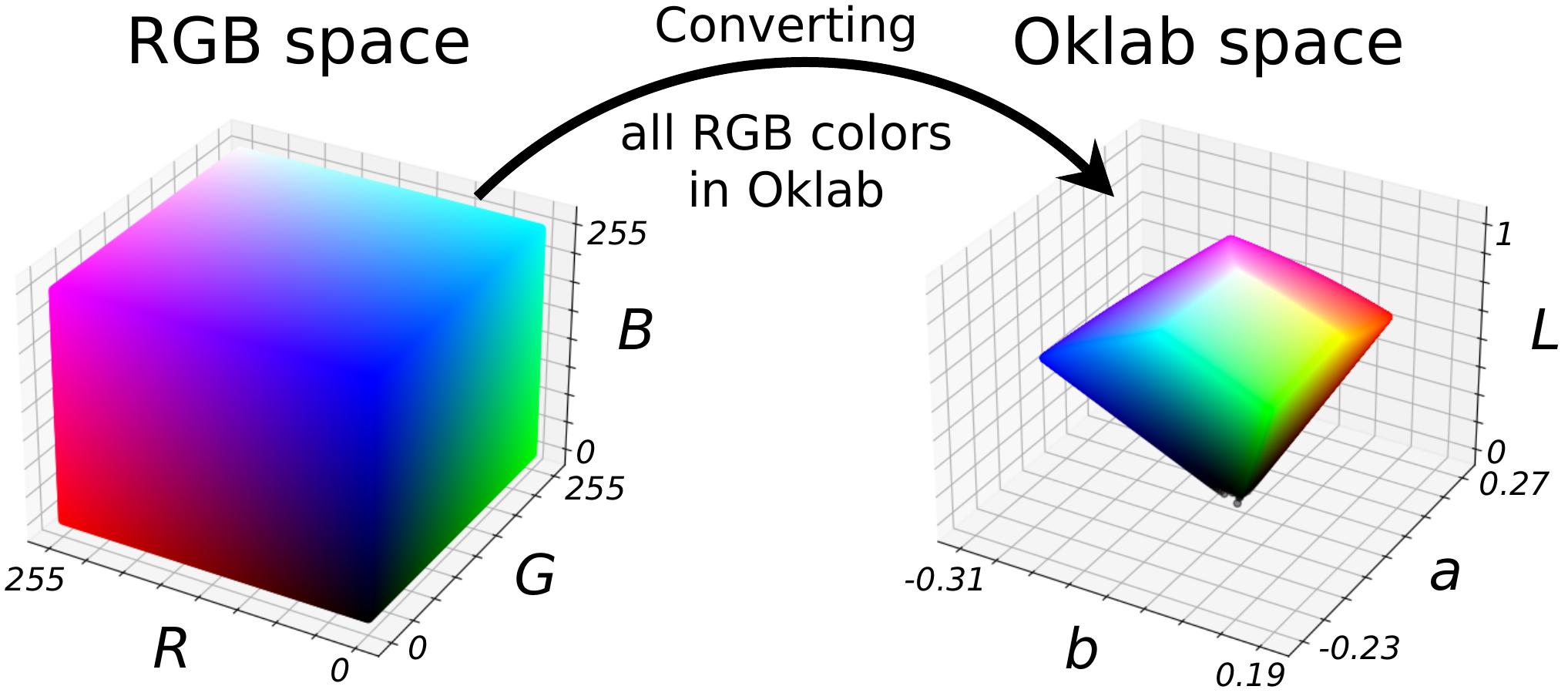}
\caption{Displacement  of all $256^3$ possible RGB color combinations in the Oklab perceptual color space.}
\label{color-conv}
\end{figure}

Oklab has the same color representation structure as CIE Lab, a color is represented by the three components (\textit{L}, \textit{a}, \textit{b}) where: \textit{L} represents the perceived brightness in the range of 0 (black) and 1 (white); while \textit{a} and \textit{b} are the axes of the opposite space \cite{hering1964outlines}, they are theoretically unbounded, but in practice, if we convert from the RGB color space, where the values are discrete and finite, as shown in Figure \ref{color-conv}, the \textit{a} axis goes in the range of -$0.23$ (green) to +$0.27$ (light magenta) and the \textit{b} axis ranges from -$0.31$ (blue) to +$0.19$ (yellow).
The properties of perceptually uniform and linear hue permit the calculation of the perceptual distance $\Delta E$ between two color points $p_1 = (L_1, a_1, b_1)$ and $p_2 = (L_2, a_2, b_2)$ in Oklab space via the straightforward Euclidean distance between the components of $p_1$ and $p_2$.

\begin{equation}
    \Delta E(p_1, p_2) = \sqrt{(L_1-L_2)^2 + (a_1-a_2)^2 + (b_1-b_2)^2}
\end{equation}
In our notion of perceptual color dissimilarity, we extend this distance to all the pixels of the two images that are the subject of the comparison. Let X and Y be two RGB images of size $M \times N \times 3$, we convert both images to the Oklab color space obtaining $\hat{X}$ and $\hat{Y}$ respectively, and then apply the following dissimilarity equation:

\begin{equation}
    OK(\hat{X}, \hat{Y}) = \frac{\sum_{i=1}^M \sum^{N}_{j=1} \Delta E(\hat{X}_{i,j}, \hat{Y}_{i,j})}{MN}
\end{equation}
where $\hat{X}_{i,j}$ and $\hat{Y}_{i,j}$ are two pixels in the Oklab color space of image $X$ and $Y$ respectively; distances in Oklab space fall in the range between $0$ and $1$, where $0$ corresponds to perfect equality and $1$ to maximum dissimilarity between the colors.

\subsection{EDOKS}\label{sec:edoks}

In addition to the dissimilarity terms, we also propose here a single global similarity metric EDOKS, which is derived from the two dissimilarity terms EMD and OK. We suggest a single overall index that exhibits an average behavior with respect to the two proposed indices. We hypothesize that this may facilitate its utilization by users. 

The energy-based normalization process for the generation of signatures, as outlined in algorithm \ref{alg:signature}, enables signatures within the range of 0 and 1. However, it should be noted that EMD can surpass 1, as the combination with the flow matrix in certain instances can exceed 1. Conversely, the OK term is constrained within the limits of 0 and 1. 

Given that both terms function within a comparable scale, the proposed equation aims to encapsulate the mean behavior between the two terms, striving to assign equal weight to both. Given two images $X$ and $Y$ to compare, the EDOK term is calculated with the following equation:

\begin{equation}\label{eq:alpha}
        EDOK(X, Y) = \alpha EMD(S_X,S_Y) + (1-\alpha) OK(\hat{X},\hat{Y})
\end{equation}
where $S_X$ and $S_Y$ are the signatures of the $X$ and $Y$ images respectively, $\hat{X}$ and $\hat{Y}$ are the $X$ and $Y$ images in the Oklab color space respectively, and $\alpha$ is a weight between 0 and 1 that is used to weigh the two terms based on the context, so that greater priority can be given to the appropriate term.

Considering that EDOK continues to be a dissimilarity term, we present the term EDOKS which is the similarity term obtained as the reciprocal of EDOK:
\begin{table*}[tb]
\centering
\caption{IQA metrics execution times when comparing, at different resolutions, two images of the LIUK4-v2 Dataset distorted with Gaussian Blur. Tests performed on Intel\textsuperscript{\tiny{\textregistered}} Xeon\textsuperscript{\tiny{\textregistered}} Gold 5215 CPU. The reported times are in sec.}
\label{tab:liuk}
\resizebox{\textwidth}{!}{
\begin{tabular}{lcccccccccccccccccccc}
\toprule
\textbf{Resolution} & \textbf{EDOKS} & \textbf{PSNR} & \textbf{SSIM} & \textbf{GMSD} & \textbf{MS-GMSD} & \textbf{MDSI} & \textbf{VIF} & \textbf{PSNR-HVS-M} & \textbf{IW-SSIM} & \textbf{MS-SSIM} & \textbf{FSIM} & \textbf{VSI} & \textbf{HaarPSI} & \textbf{DSS} & \textbf{SRSIM} & \textbf{LPIPS} & \textbf{PIEAPP} & \textbf{DISTS} & \textbf{TOPIQ} \\
\midrule
480p &
0.351 & 0.005 & 0.112 & 0.020 & 0.031 &
0.025 & 1.166 & 0.126 & 0.726 & 1.486 &
0.093 & 0.057 & 0.039 & 0.048 & 0.027 &
0.042 & 2.966 & 0.979 & 0.502 \\

720p &
1.219 & 0.013 & 0.306 & 0.021 & 0.043 &
0.030 & 2.562 & 0.301 & 1.542 & 3.494 &
0.175 & 0.114 & 0.096 & 0.079 & 0.037 &
0.065 & 7.164 & 2.385 & 1.423 \\

1080p &
5.951 & 0.066 & 0.709 & 0.166 & 0.297 &
0.191 & 5.940 & 0.582 & 3.902 & 8.364 &
0.282 & 0.387 & 0.361 & 0.265 & 0.176 &
0.153 & 16.159 & 5.085 & 3.182 \\
\bottomrule
\end{tabular}}
\end{table*}
\begin{equation} \label{eq:edoks}
        EDOKS(X, Y) = \frac{1}{EDOK(X, Y) + c}
\end{equation}
where $c$ is a very small constant to avoid performing a division by zero, which might happen when $X$ and $Y$ are equal; because when images are equal EDOK returns 0 as the dissimilarity score.

Although EDOKS is a single performance score, we suggest not only to use it, but also to evaluate the EMD and OK indices at the same time, in order to evaluate in detail the quality of textures and colors, which would not be possible with the average EDOKS overall.

\subsection{Computational Analysis}

A fundamental aspect that needs to be clarified, given the modularity of our metric, is the computational complexity of EDOKS. 

Assuming that each image has $P$ pixels, the most expensive operations of the Oklab term are the conversion to another color space, which requires $O(P)$, and the calculation of the similarity distance for each pair of pixels, which has a cost of $O(P)$. 

Regarding the application of the Gabor filter bank, each filter is applied with a convolution that has a cost of $O(PG^2)$, where $G$ is the size of the Gabor kernel. As the number of filters increases, the complexity of this operation remains unchanged. This is because the application of the 24 Gabor filters has been parallelized across multiple CPU or GPU cores. The subsequent clustering operation has a cost of $O(l^3)$, where $l$ is the number of patches in the image.

Finally, for EMD we used the implementation of \cite{rubner_earth_2000}, where the most expensive operation is the loop to find the optimal solution, which has a computational cost of $O(K^2)$, where $K$ is the size of the two signatures corresponding to the number of clusters, which generally limits the number of elements. 

Although some terms have high asymptotic computational complexity, in practical terms it should be noted that the calculation of the EDOKS value is usually applied to images with limited dimensionality. As shown in Table \ref{tab:liuk}, we calculated the times on one image of the Large-scale Ideal Ultra high definition 4K version 2 (LIUK4-v2) dataset \cite{Liu4K} at different resolutions compared to its Gaussian blurred version, obtaining a runtime consistent with other SOTA IQA metrics. 

As shown by the results obtained, it can be seen that models considered to be more performant, such as deep architectures, require longer execution times.

\section{Results}

To evaluate the robustness of our metric to human perception, we conducted experiments on datasets in which the similarity between images is assessed by humans, as in BAPPS datasets \cite{zhang_unreasonable_2018}. We used the Just Noticeable Difference (JND) and Two Alternative Forced Choice (2AFC) subsets, contained in BAPPS, to conduct our experiments. We show that our metric is closer to human perception than the SOTA metrics. In our experiments, we chose not to use popular datasets such as LIVE \cite{sheikh_statistical_2006}, TID2008 \cite{ponomarenko2009tid2008}, CSIQ \cite{chandler_most_2010}, and TID2013 \cite{ponomarenko_image_2015}.  This is because, as demonstrated in \cite{zhang_unreasonable_2018}, they are outdated and contain a few types of distortions.

For all experiments performed in this section, the EDOKS parameters were initialized as follows: $p = 128$ to achieve a good compromise in the extraction of regional patterns, $\alpha = 0.5$ to obtain the same contribution from both terms, and $c$ as the minimum float value representable in Python, which provides an excessive contribution to the equation \ref{eq:edoks} and prevents division by zero.

\begin{figure}[t]
\includegraphics[width=\linewidth]{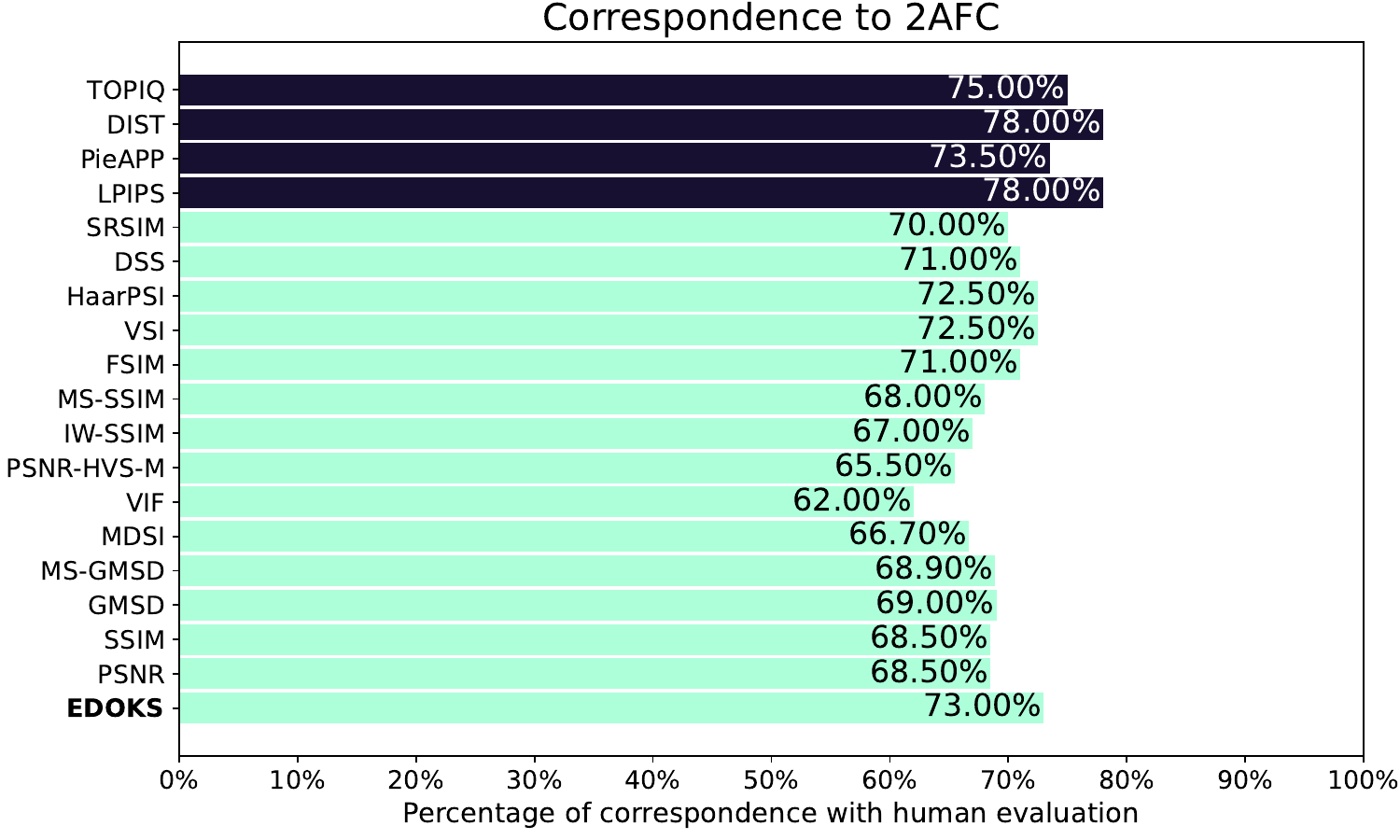}
\caption{Correspondences of similarity metrics with the 2AFC dataset. The horizontal axis represents the perceptual similarity of humans who labeled the 2AFC dataset, as reported in \cite{zhang_unreasonable_2018}. The results show that, compared with other low-level FR methods, EDOKS is the closest similarity metric to human perceptual similarity, while it is close to the results of other deep metrics.}
\label{chart2afc}
\end{figure}
\begin{figure*}[b]
\includegraphics[width=\linewidth]{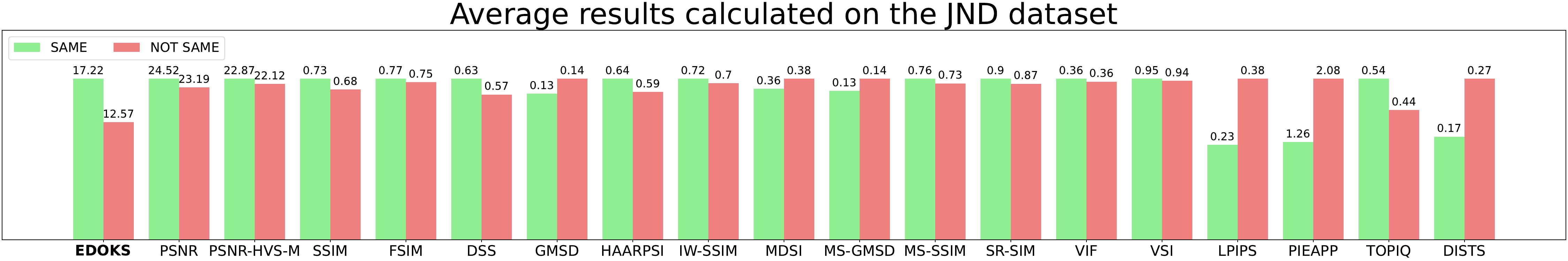}
\caption{Average scores responses obtained by EDOKS and the SOTA metrics calculated on the JND dataset at the pairs of images rated same (\textcolor{lightgreen}{green}) and not same (\textcolor{lightred}{red}) by humans. The discrepancy in scores between same and not same subsets obtained by our EDOKS metric shows behavior more consistent with human perception than the scores obtained by other low level metrics, while it is close to the results of other deep metrics.}
\label{chartJND}
\end{figure*}

\subsection{Datasets}
Our goal is to evaluate perceptual similarity, to do so we need a dataset that contains a significant number of distortions that are as varied as possible. We therefore opted to use the BAPPS dataset as it contains countless distortions, including artifacts generated by deep CNN models and geometric distortions. BAPPS is patch-oriented, with images distorted at the patch level to appreciate local distortions.

We used the validation subset containing images obtained from the RAISE1k dataset \cite{dang-nguyen_raise:_2015} and including both traditional and CNN-based distortions.

The 2AFC dataset \cite{zhang_unreasonable_2018} contains the assessments of a 5-person jury. Each sample submitted to the jury consists of a patch $x$ derived from an original image of the RAISE1k dataset and two distorted versions $x_1$ and $x_2$ of the same. The jury had to decide which of the two distorted patches, $x_1$ or $x_2$, was perceptually most similar to the original $x$ patch.

The JND dataset \cite{zhang_unreasonable_2018} was created to be more objective than the 2AFC dataset. The three-person jury had less time to look at the images and provide a judgment. Only two images are shown to create this dataset: the reference image and a distorted version of it. The jury is then asked whether the images are the \textit{same} or \textit{not same}. The two images are shown for 1 second each, with an interval of 250 ms between them, so that the jury can answer immediately based solely on their perception without being biased.

\subsection{Accuracy}
Using the human-labeled 2AFC and JND datasets, we evaluated the similarity of the samples with our EDOKS metric. We compared its values to those obtained with the state-of-the-art (SOTA) metrics to assess which metric best aligns with human perception.

Specifically, since each sample in the 2AFC dataset is a triplet ($x$, $x_1$, $x_2$), where the jury evaluated which of $x_1$ and $x_2$ is closer to $x$, we also used our metric in our experiments to evaluate the similarity between $x_1$ and $x$ and between $x_2$ and $x$ in order to determine which of the two pairs is more similar. Qualitative comparisons were calculated as in \cite{zhang_unreasonable_2018} and, in Figure \ref{chart2afc}, the histogram shows how many times the similarity metrics agree with the decisions of the jury that evaluated the 2AFC dataset. The results show that our metric has more perceptual behavior than other low-level metrics, while it is close to the results of other deep metrics. Our goal was not to obtain a total match with human perception, but to show that our metric is more perceptual than the others.

We exploited JND to assess how much the metric penalizes or perceptually rewards similarity between images. Therefore, from the entire JND dataset, we performed a pre-processing by extracting two subsets: \textit{same} and \textit{not same}. The subset \textit{same} contained all pairs of images in which the jury unanimously agrees that the pairs of images have a high level of similarity. The second subset, \textit{not same}, contained all image pairs in which the jury unanimously agrees in rating the images in each pair as different. From the JND dataset, therefore, all pairs in which the jury's opinions were discordant were excluded. As shown in Figure \ref{chartJND}, we averaged the scores of the EDOKS and the SOTA metrics over the two subsets; remember that when the \textit{same} subset has a lower average value than the \textit{not same} subset, it is because the metric evaluated is a dissimilarity metric. 

This experiment was useful to be able to assess how much the scores of the metrics differ by changing subsets. It is found that EDOKS clearly distinguishes \textit{same} image pairs from \textit{not same} image pairs, assigning on average very different scores. 

Moreover the low level metrics show difficulty in perceptually distinguishing the two subsets, while deep metrics can clearly distinguish between the two subsets. 

Furthermore, this is a good indicator of what value EDOKS should take when two images in this dataset are similar.
\subsection{Statistical comparison}

For the statistical analysis of our similarity metric, a comparison with MOS is required, since the dataset used does not directly provide MOS values. We obtained them implicitly from the votes of the jury that labeled the dataset. 

Specifically, our MOS scale includes four possible rating levels, where 0 indicates that no judge considers two images to be the same, while 1 indicates that all judges agreed on the similarity of two images.

To evaluated the performance of image quality metrics, we used three
correlation measures: Spearman Rank Order Correlation Coefficient
(SROCC), Kendall Rank Order Correlation Coefficient (KROCC), and Pearson
Linear Correlation Coefficient (PLCC). 

SROCC and KROCC can measured the prediction monotonicity of a metric, while the PLCC measured the linear relationship between predicted and reference scores. To compute the PLCC we applied a nonlinear regression analysis, we used the following mapping function as suggested in \cite{sheikh_statistical_2006}:

\begin{equation}
f(x) =  \beta_1 \Big( \frac{1}{2} - \frac{1}{1 + e^{\beta_2 (x - \beta_3)}} \Big) + \beta_4 x + \beta_5
\end{equation}
where $\beta_i , i = 1, 2,..., 5$, are parameters to be fitted.

For all these three coefficients, the higher the value, the better the correlation.

\begin{table}[t]
\centering
\caption{Overall Performances of Low-Level IQA Indices on the JND Dataset}
\label{tab:low_stat}
\begin{tabular}{lccc}
\toprule
\textbf{Metric} & \textbf{SROCC}$\raisebox{0.2ex}{$\uparrow$}$ & \textbf{KROCC}$\raisebox{0.2ex}{$\uparrow$}$ & \textbf{PLCC}$\raisebox{0.2ex}{$\uparrow$}$ \\
\midrule
\textbf{EDOKS} & \textbf{0.537} & \textbf{0.419} & \textbf{0.548} \\
PSNR & 0.390 & 0.304 & 0.465 \\
SSIM & 0.365 & 0.284 & 0.405 \\
FSIM & 0.343 & 0.269 & 0.479 \\
DSS & 0.427 & 0.334 & 0.435 \\
GMSD & 0.370 & 0.290 & 0.403 \\
HaarPSI & 0.448 & 0.353 & 0.519 \\
IW-SSIM & 0.415 & 0.327 & 0.460 \\
MDSI & 0.377 & 0.294 & 0.427 \\
MS-GMSD & 0.364 & 0.285 & 0.406 \\
MS-SSIM & 0.402 & 0.317 & 0.474 \\
PSNR-HVS\-M & 0.335 & 0.261 & 0.426 \\
SR-SIM & 0.429 & 0.334 & 0.432 \\
VIF & 0.294 & 0.230 & 0.352 \\
VSI & 0.458 & 0.360 & 0.519 \\
\bottomrule
\end{tabular}
\end{table}

\begin{table}[b]
\centering
\caption{Overall Performances of EDOKS and Deep IQA Indices on the JND Dataset}
\label{tab:deep_stat}
\begin{tabular}{lccc}
\toprule
\textbf{Metric} & \textbf{SROCC}$\raisebox{0.2ex}{$\uparrow$}$ & \textbf{KROCC}$\raisebox{0.2ex}{$\uparrow$}$ & \textbf{PLCC}$\raisebox{0.2ex}{$\uparrow$}$ \\
\midrule
\textbf{EDOKS} & 0.537 & 0.419 & 0.548 \\
LPIPS & 0.633 & 0.499 & 0.640 \\
PIEAPP & 0.551 & 0.432 & 0.550 \\
DISTS & \textbf{0.649} & \textbf{0.513} & \textbf{0.671} \\
TOPIQ & 0.585 & 0.461 & 0.614 \\
\bottomrule
\end{tabular}
\end{table}

Table \ref{tab:low_stat} shows that our EDOKS metric significantly outperforms the SOTA low-level metrics. Meanwhile, Table \ref{tab:deep_stat} shows that EDOKS is close to deep metrics, which, although more accurate, lack transparency.

\begin{figure}[!t]
\includegraphics[width=\linewidth]{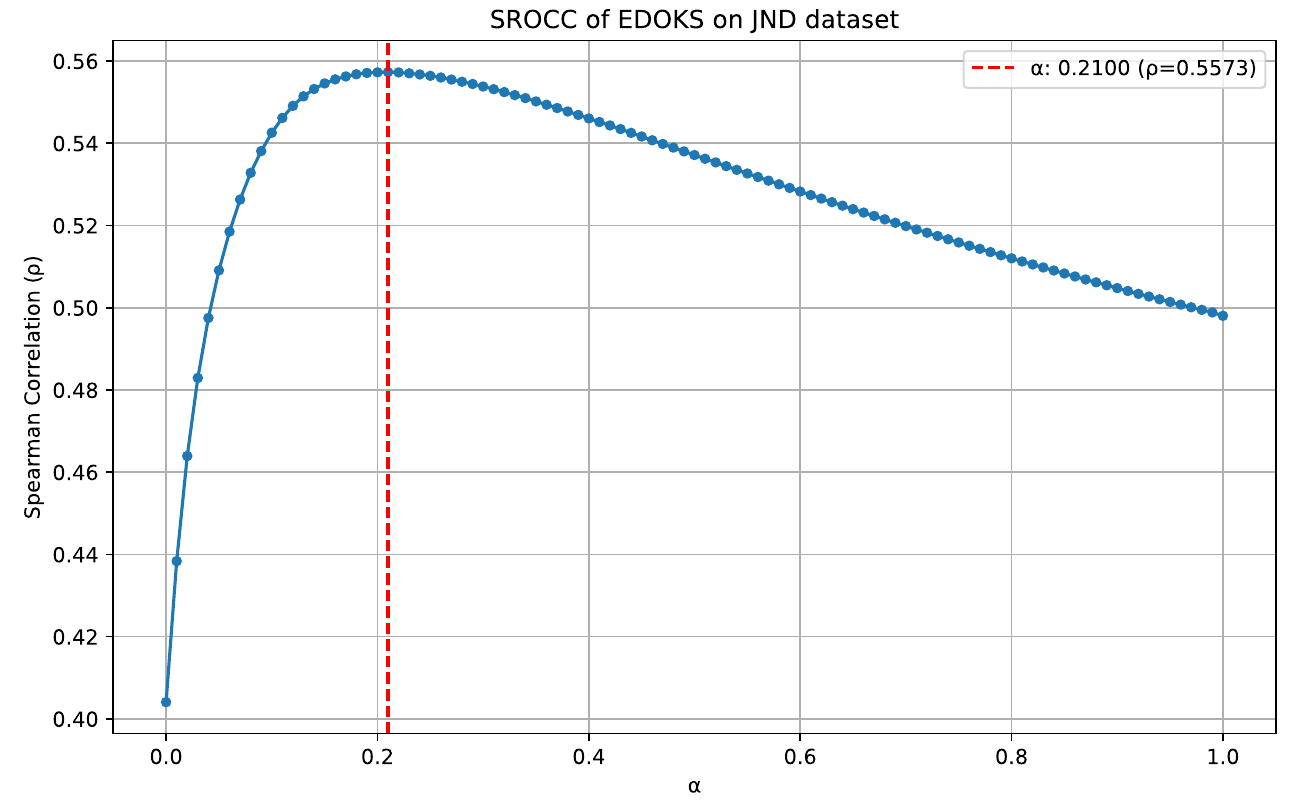}
\caption{
SROCC value as the alpha weight of Eq. \ref{eq:alpha} varies. The scores were calculated on the JND subset, and the optimal value in this case is $\alpha = 0.21$. The SROCC values were calculated by varying $\alpha$ from $0$ to $1$ in increments of $0.01$.}
\label{fig:alpha_jnd}
\end{figure} 

\subsection{Ablation Study}

In this subsection, we seek to analyze the impact of the combination of our two terms on the final results of EDOKS. The necessity to integrate a term that assesses textures and a term that evaluates colors is discussed.

In the course of our experimental investigations, we eliminated the contributions of the EMD term and the OK term by appropriately adjusting the weight parameter, designated as $\alpha$, within the Eq. \ref{eq:alpha}. The tests conducted in this chapter are analogous to those in the preceding chapter, with the distinction that the individual terms are compared with the combination of the two.

In Figure \ref{fig:alpha_jnd}, we can see the usefulness of the coefficient $\alpha$. As $\alpha$ changes, the weights of the individual terms in the equation change and so do the contributions they make to the final score. To illustrate this phenomenon, we computed the SROCC on the JND dataset as the parameter $\alpha$ varies. It can be seen that with $\alpha$ tending towards 0, the absence or small contribution of the EMD term in the formula had a significant impact on the SROCC obtained on the JND dataset.

After showing the behavior of the metric as $\alpha$ varies, we analyze the extreme cases, in which the EDOKS score is obtained with the sole contribution of the EMD term or with the sole contribution of the OK term.

Table \ref{tab:ablation_stat} shows the scores of the various experiments, demonstrating that the combination of the two terms yields a better result than the use of the individual terms.

In general, it should be noted that distortions can occur in both colors and shapes. As demonstrated by our results, using only one of the two terms is not sufficient to evaluate both types of distortion. However, in a specific case where the type of distortion applied is known (e.g., only color distortions), we allow the possibility of configuring the weight $\alpha$ to obtain an analysis that is as consistent as possible with the case study.

\begin{table*}[h]
\centering
\caption{Study of the ablation of the edoks metric by evaluating its individual terms against itself. The AVG scores and statistical metrics (SROCC, KROCC, PLCC) were obtained from the JND dataset, while the accuracy values refer to the 2AFC dataset.}
\label{tab:ablation_stat}
\begin{tabular}{lccccccc}
\toprule
\multicolumn{1}{c}{\multirow{2}{*}{\textbf{Metric}}} & \multicolumn{3}{c}{The average scores (AVG)} & \multirow{2}{*}{\textbf{SROCC}$\raisebox{0.2ex}{$\uparrow$}$} &
\multirow{2}{*}{\textbf{KROCC}$\raisebox{0.2ex}{$\uparrow$}$} &
\multirow{2}{*}{\textbf{PLCC}$\raisebox{0.2ex}{$\uparrow$}$} &
\multirow{2}{*}{\textbf{Accuracy}$\raisebox{0.2ex}{$\uparrow$}$} \\
\cmidrule{2-4}
& \textbf{Same} & \textbf{Not Same} & \textbf{Ratio} \\
\midrule
EMD ($\alpha = 1$) & 3.70 & 2,27 & 1.63 & 0.498 & 0.388 & 0.498 & 71\%\\
OK ($\alpha = 0$) & 16.67 & 14.29 & 1.16 & 0.404 & 0.314 & 0.454 & 69\%\\
EDOKS ($\alpha = 0.5$) & 17.22 & 12.57 & 1.37 & 0.537 & 0.419 & 0.548 & 72\% \\
\bottomrule
\end{tabular}
\end{table*}

\section{Interpretability Validation}

\begin{figure*}[b]
\centering
\begin{tabular}[c]{ccccc}
    Input image & Input image & EMD difference & OK heatmap & {\small EDOKS overall map} \\
    $X$ & $Y$ & $\Delta|F|_{XY}$ & $\Delta E_{XY}$ & $\Delta O=\Delta E_{XY} \oplus \Delta|F|_{XY}$ \\
    \includegraphics[width=0.18\linewidth]{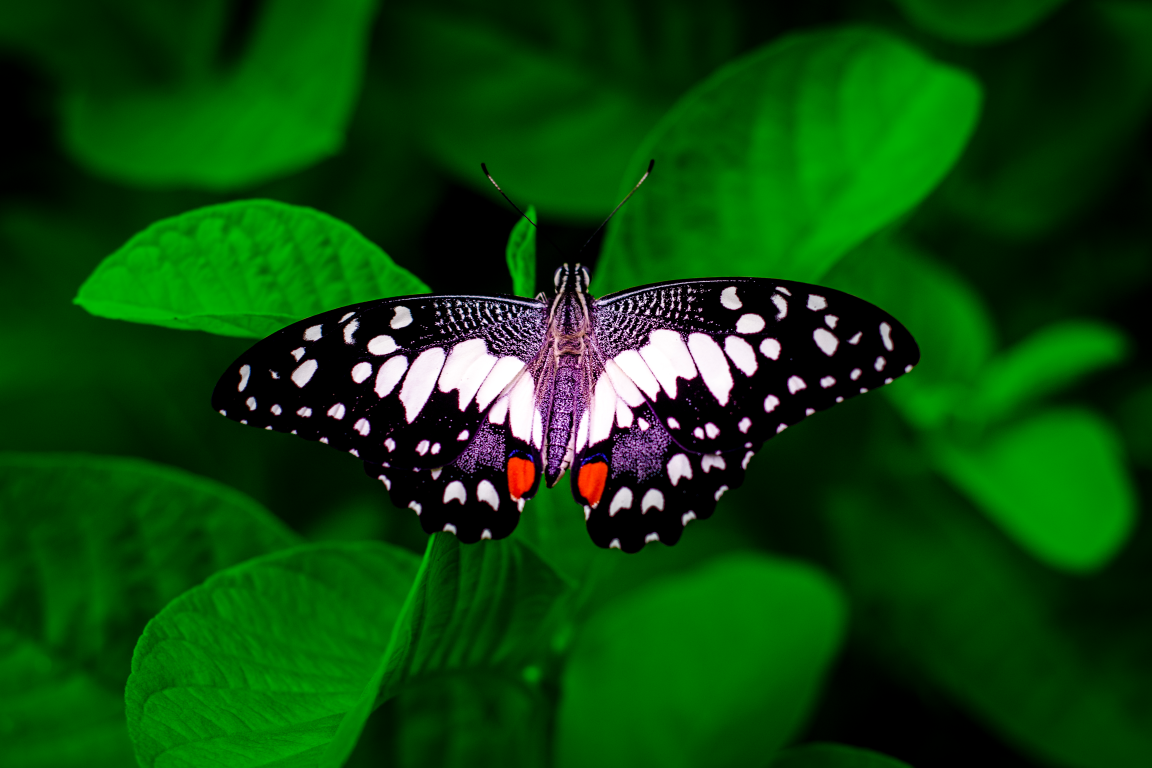} & \includegraphics[width=0.18\linewidth]{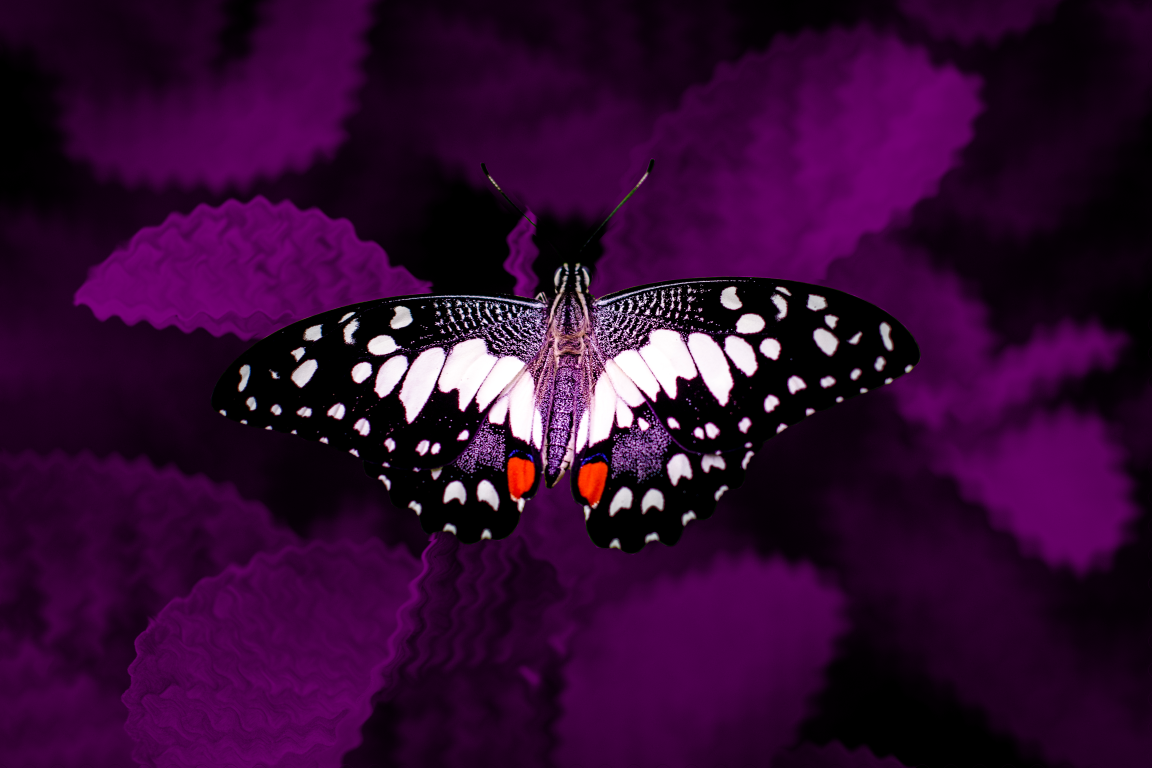}  &    \includegraphics[width=0.18\linewidth]{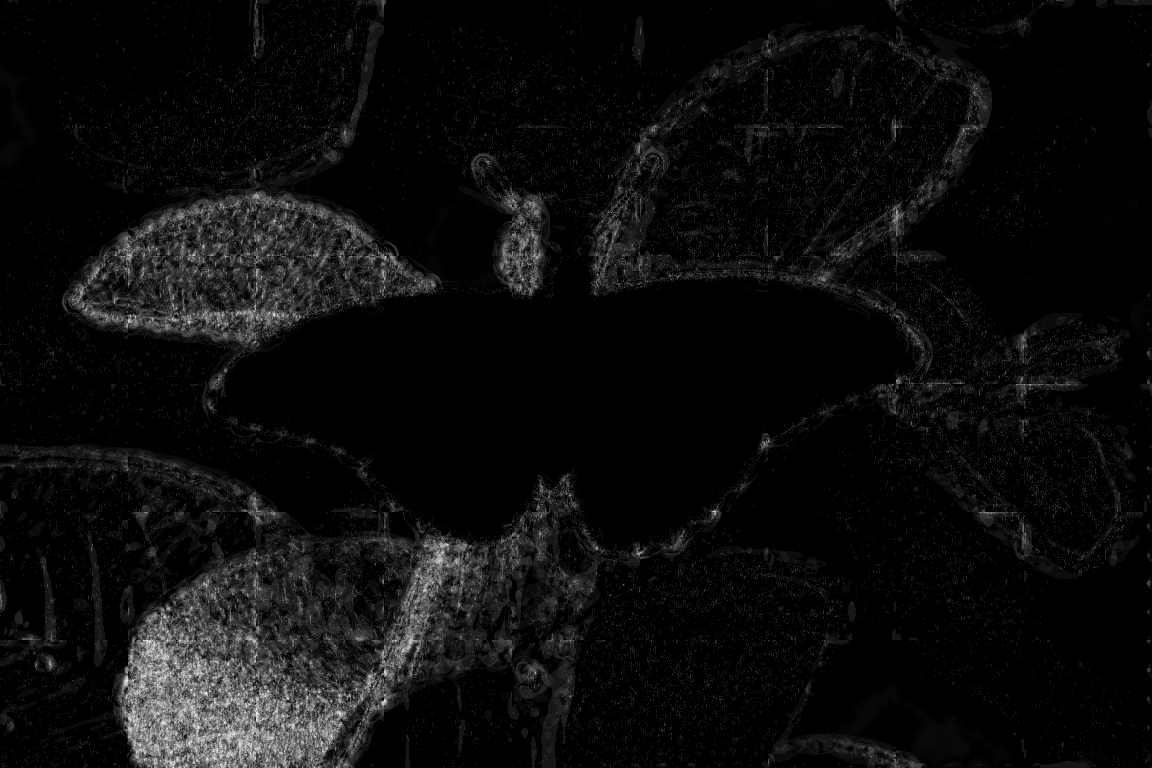} &
    \includegraphics[width=0.18\linewidth]{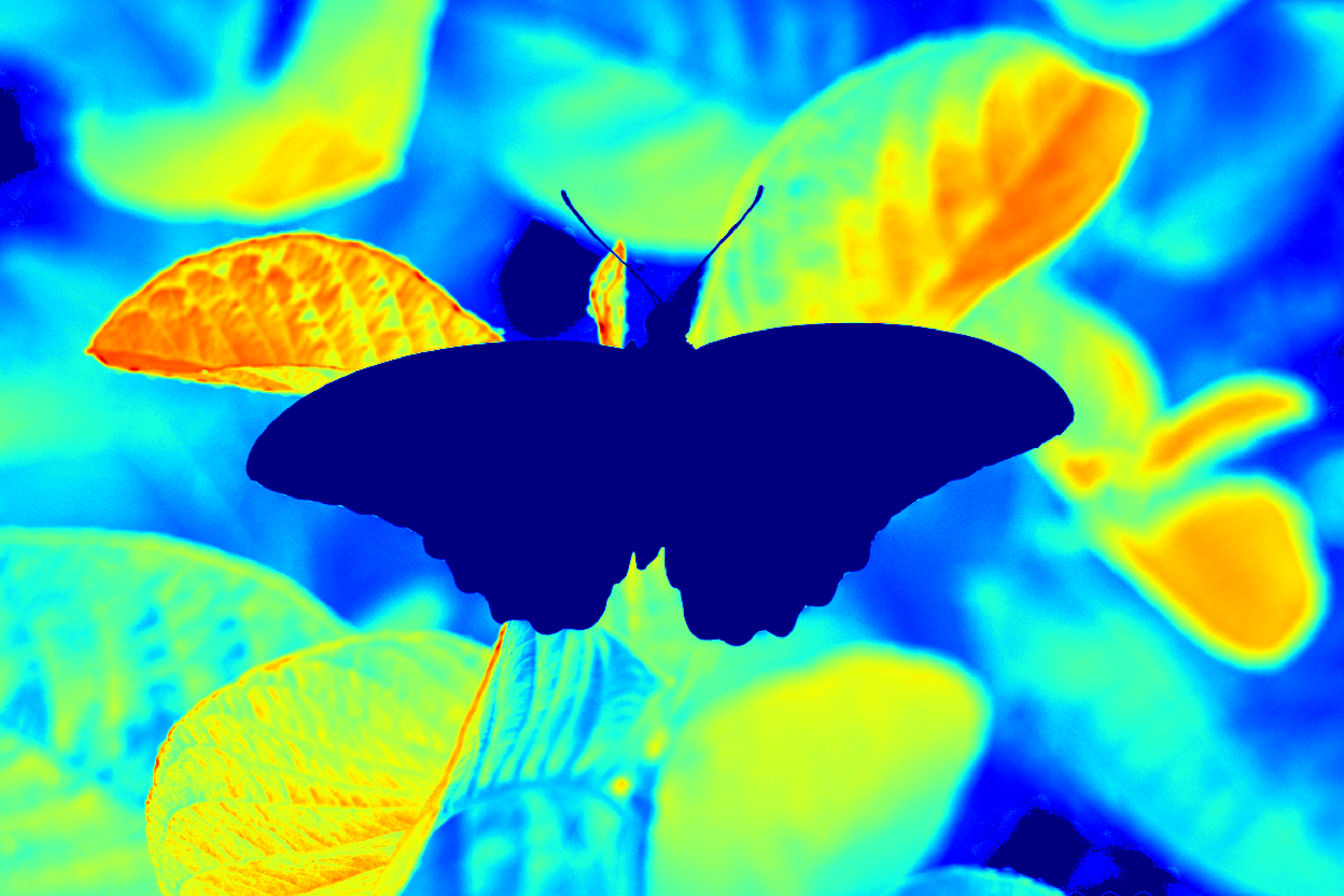} & \includegraphics[width=0.18\linewidth]{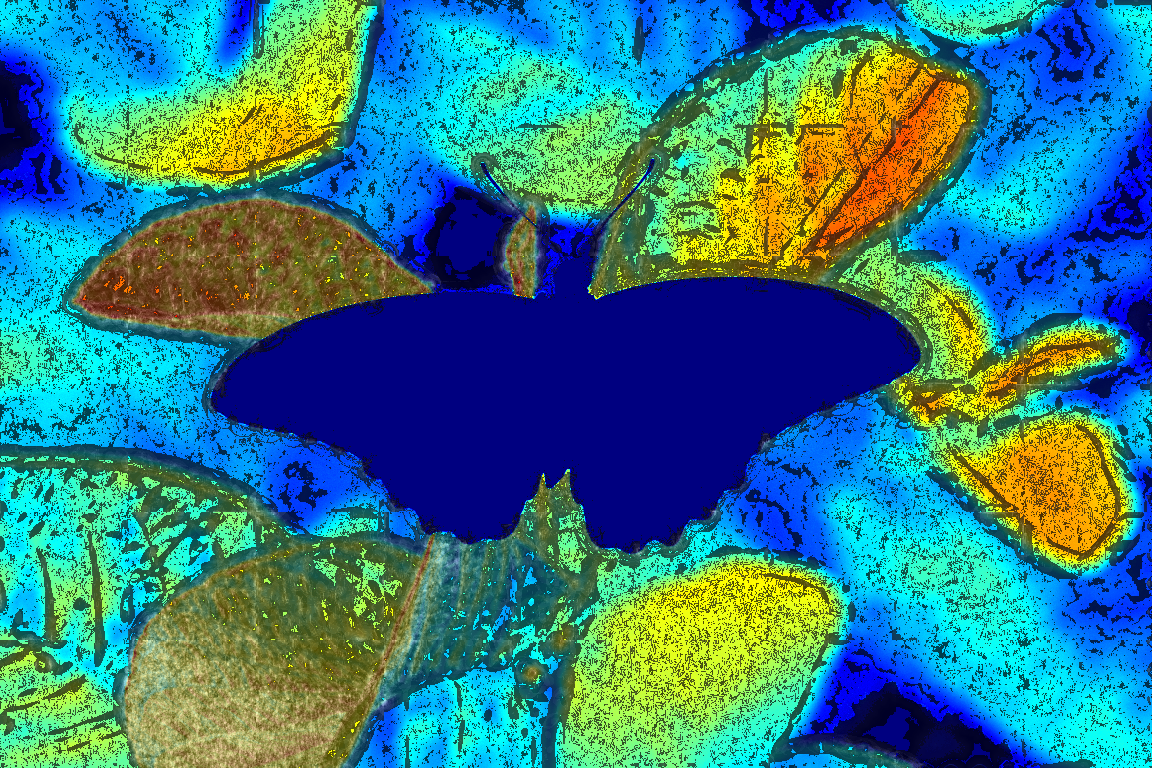} \\
    \includegraphics[width=0.18\linewidth]{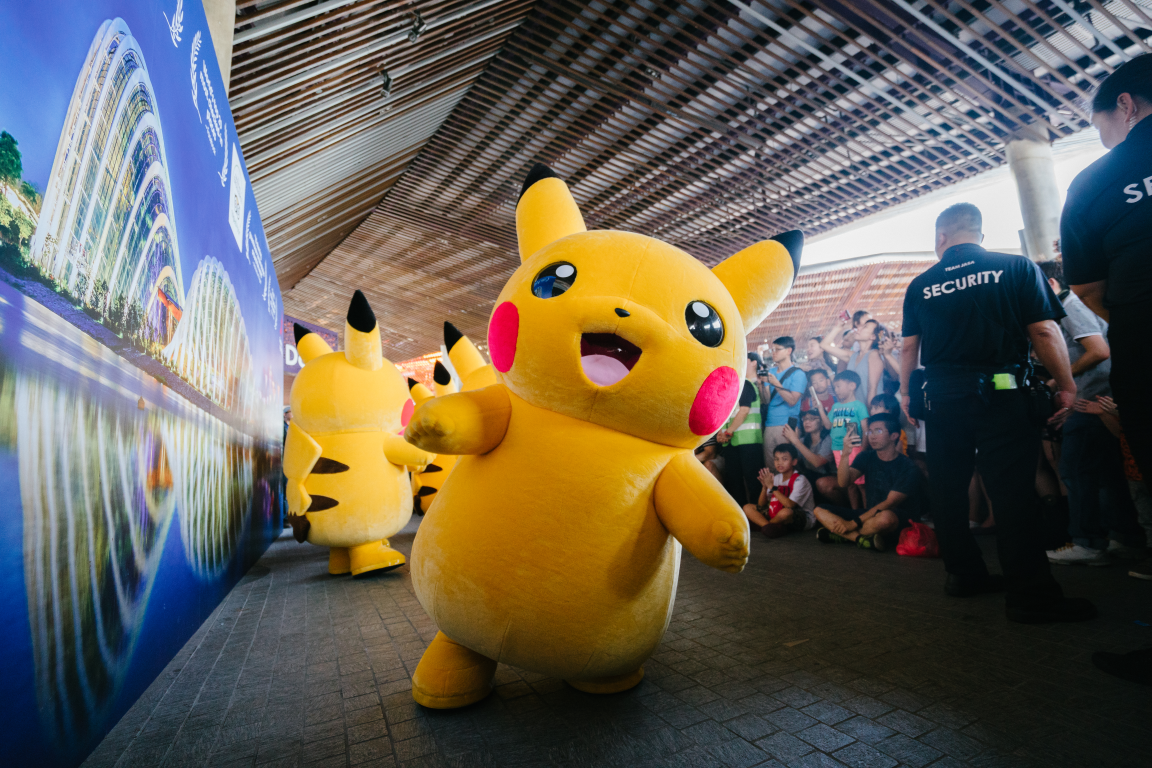} & \includegraphics[width=0.18\linewidth]{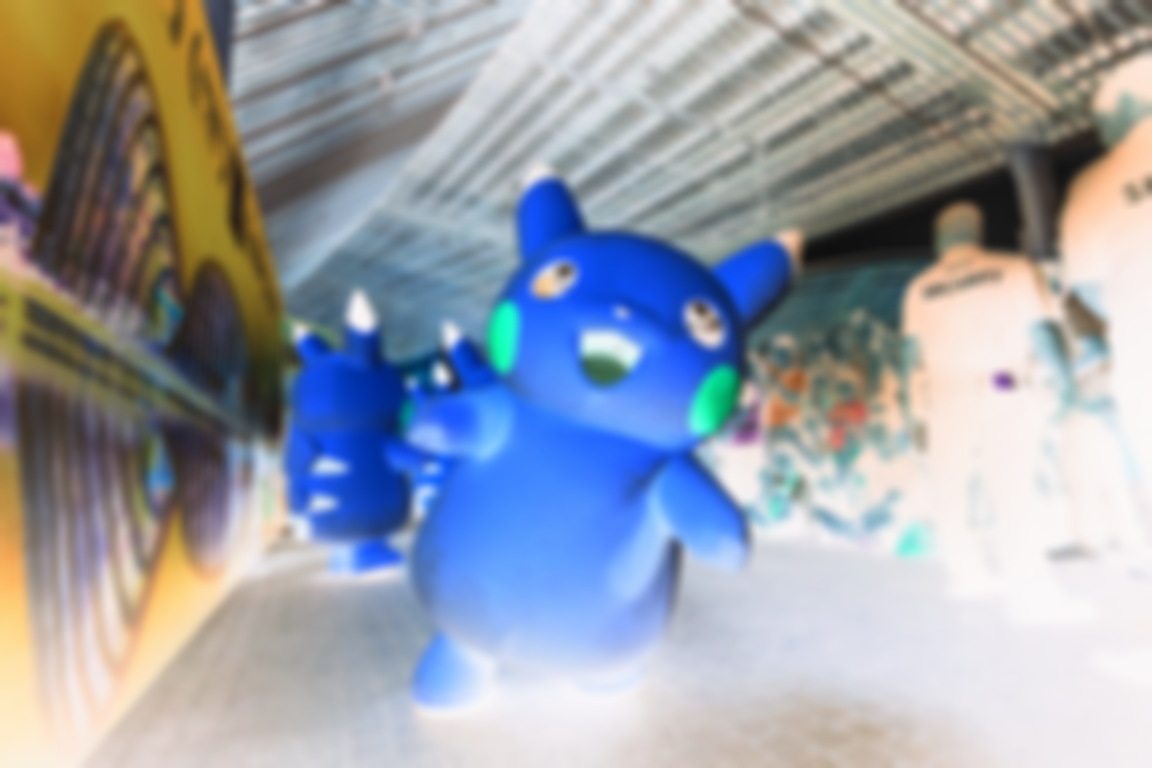}  &     \includegraphics[width=0.18\linewidth]{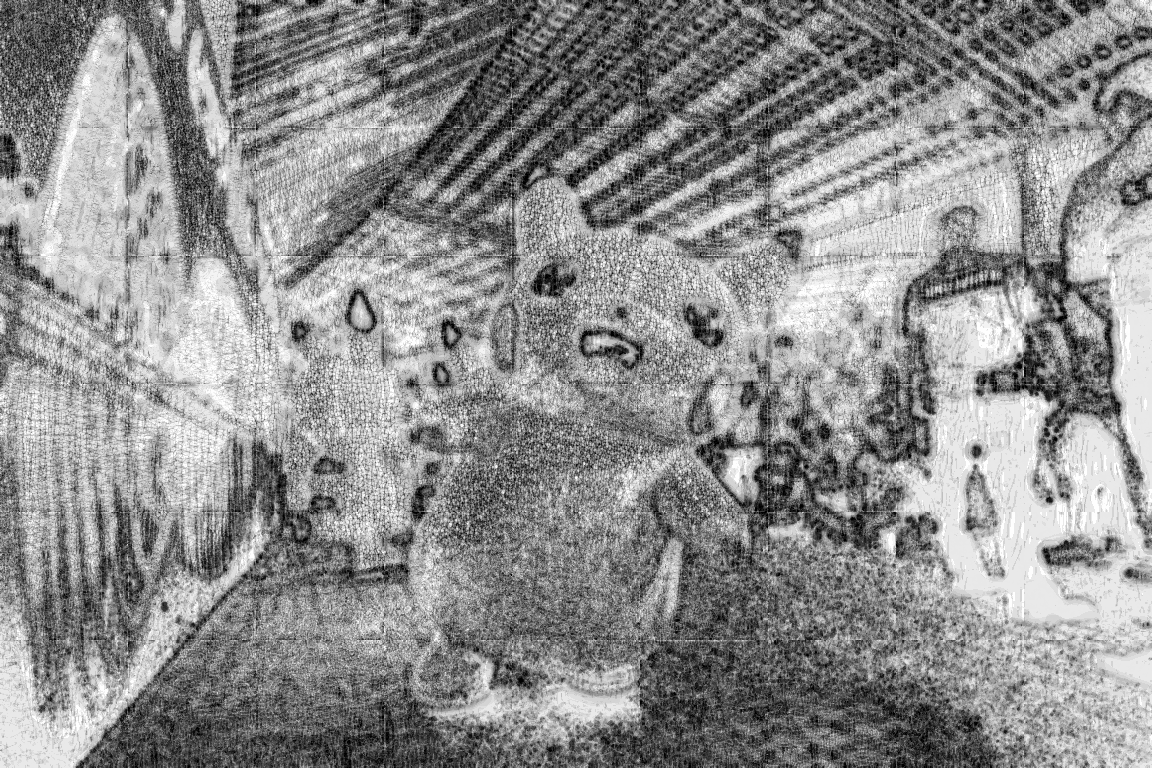} &
    \includegraphics[width=0.18\linewidth]{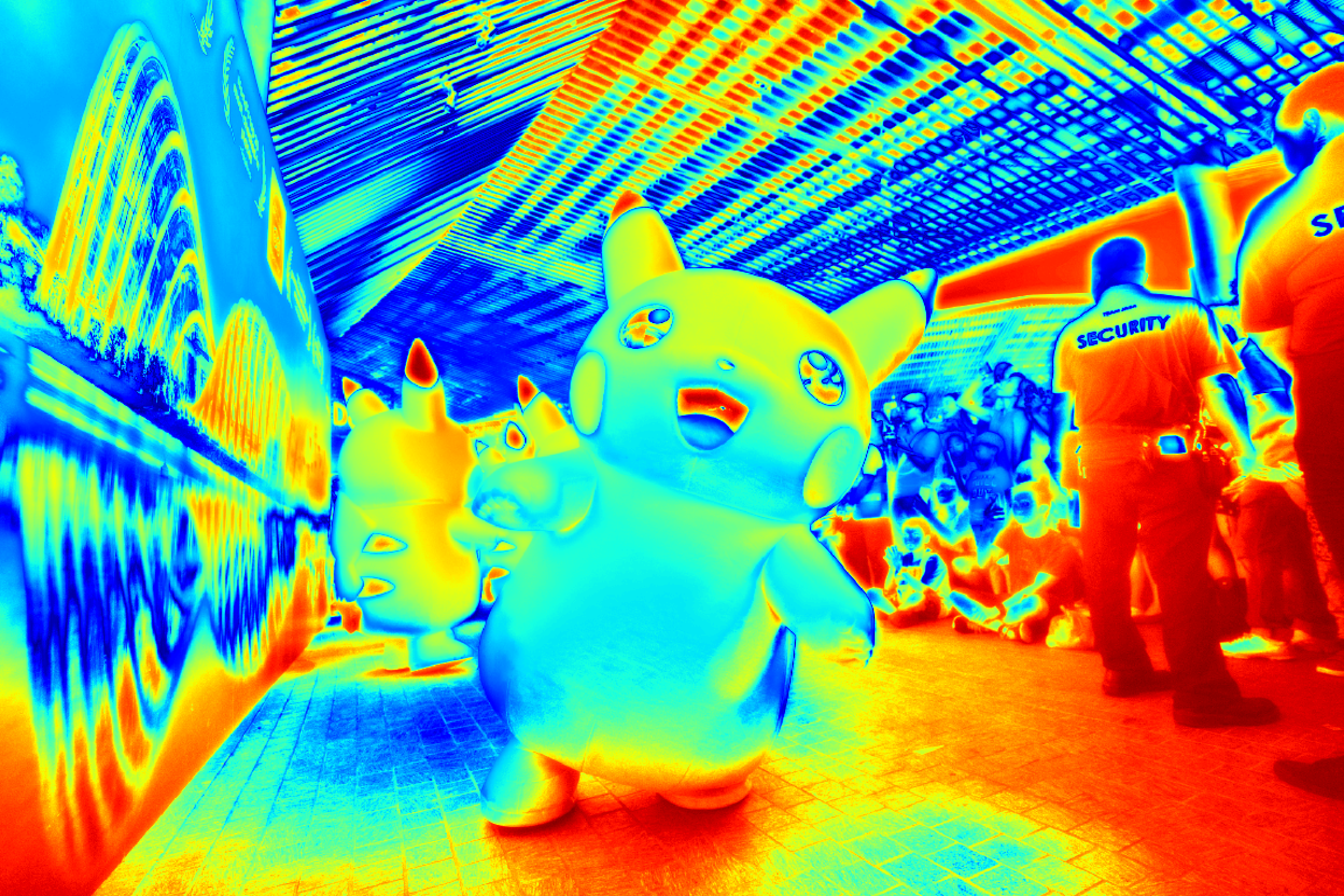} & \includegraphics[width=0.18\linewidth]{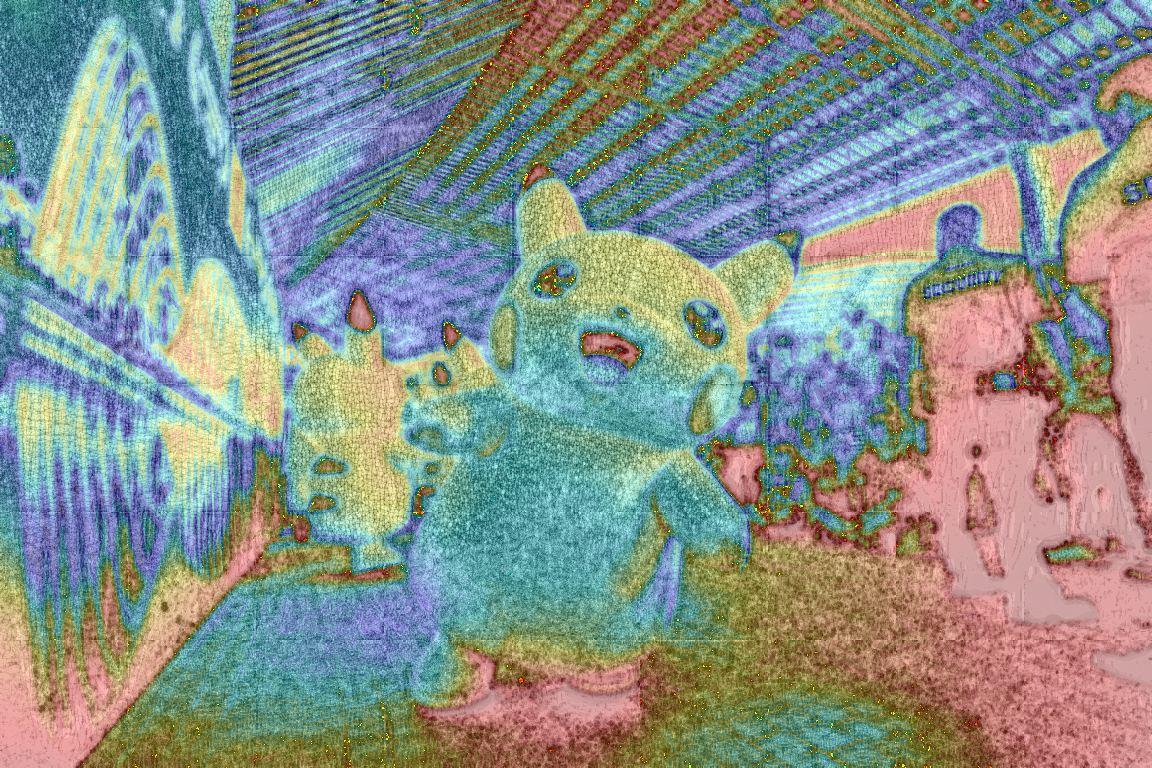} \\
    \includegraphics[width=0.18\linewidth]{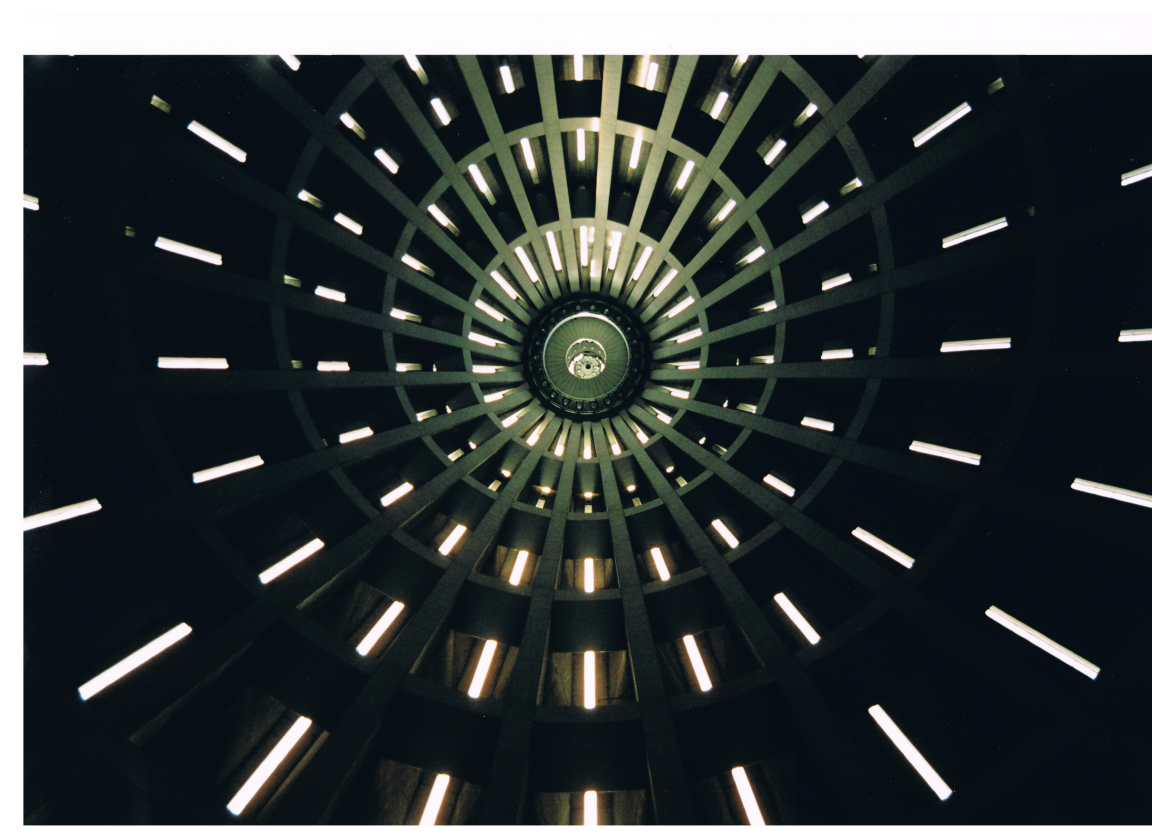} & \includegraphics[width=0.18\linewidth]{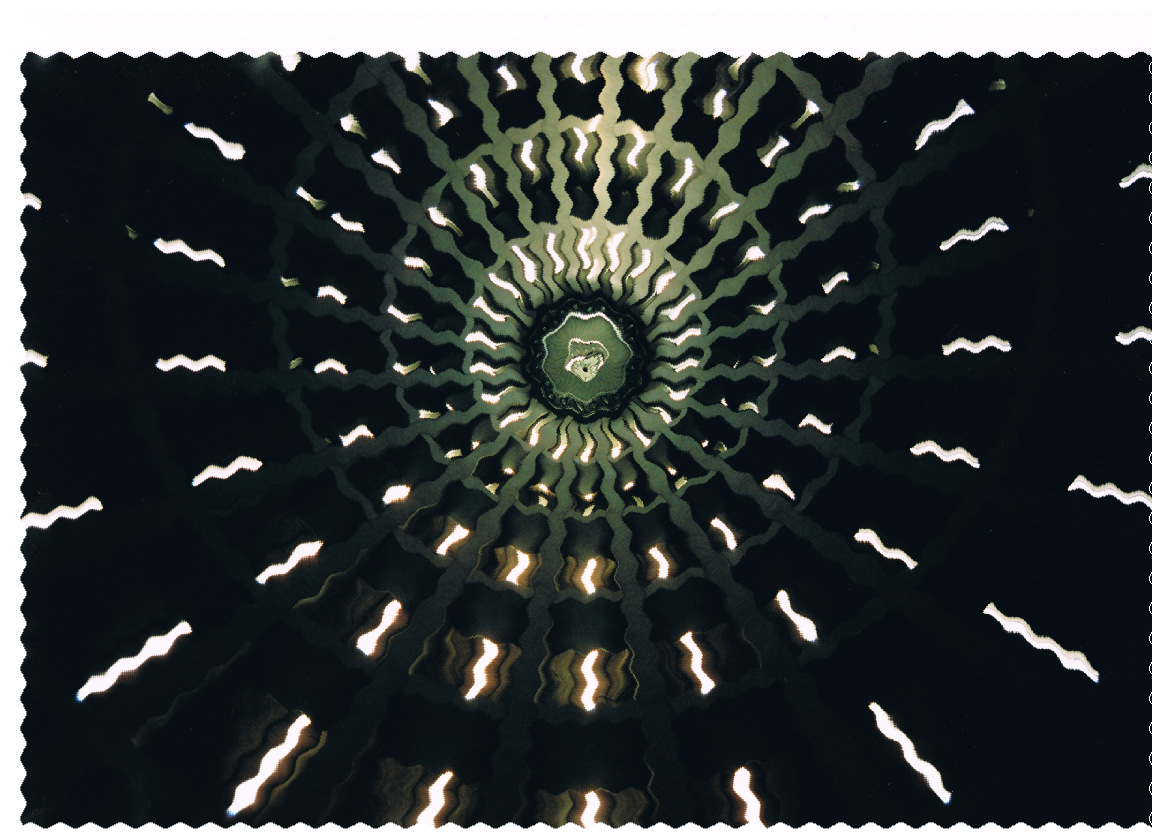}  &     \includegraphics[width=0.18\linewidth]{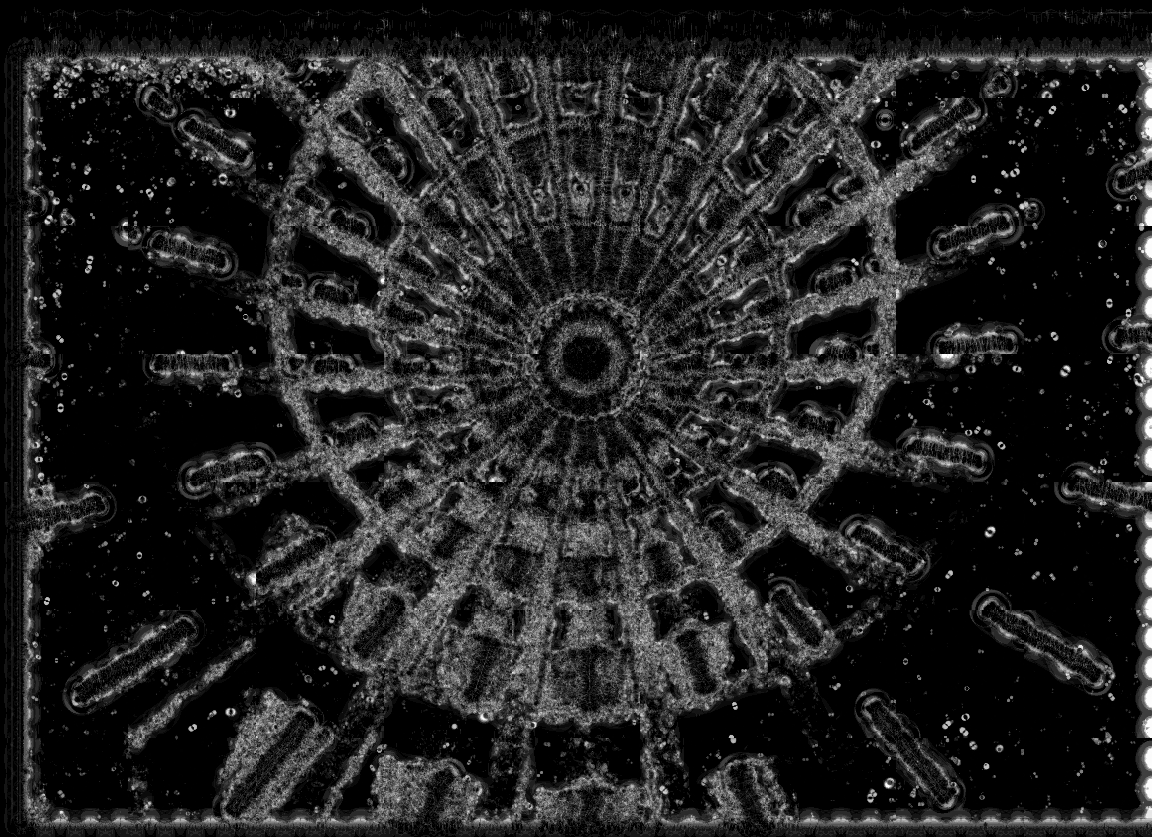} &
    \includegraphics[width=0.18\linewidth]{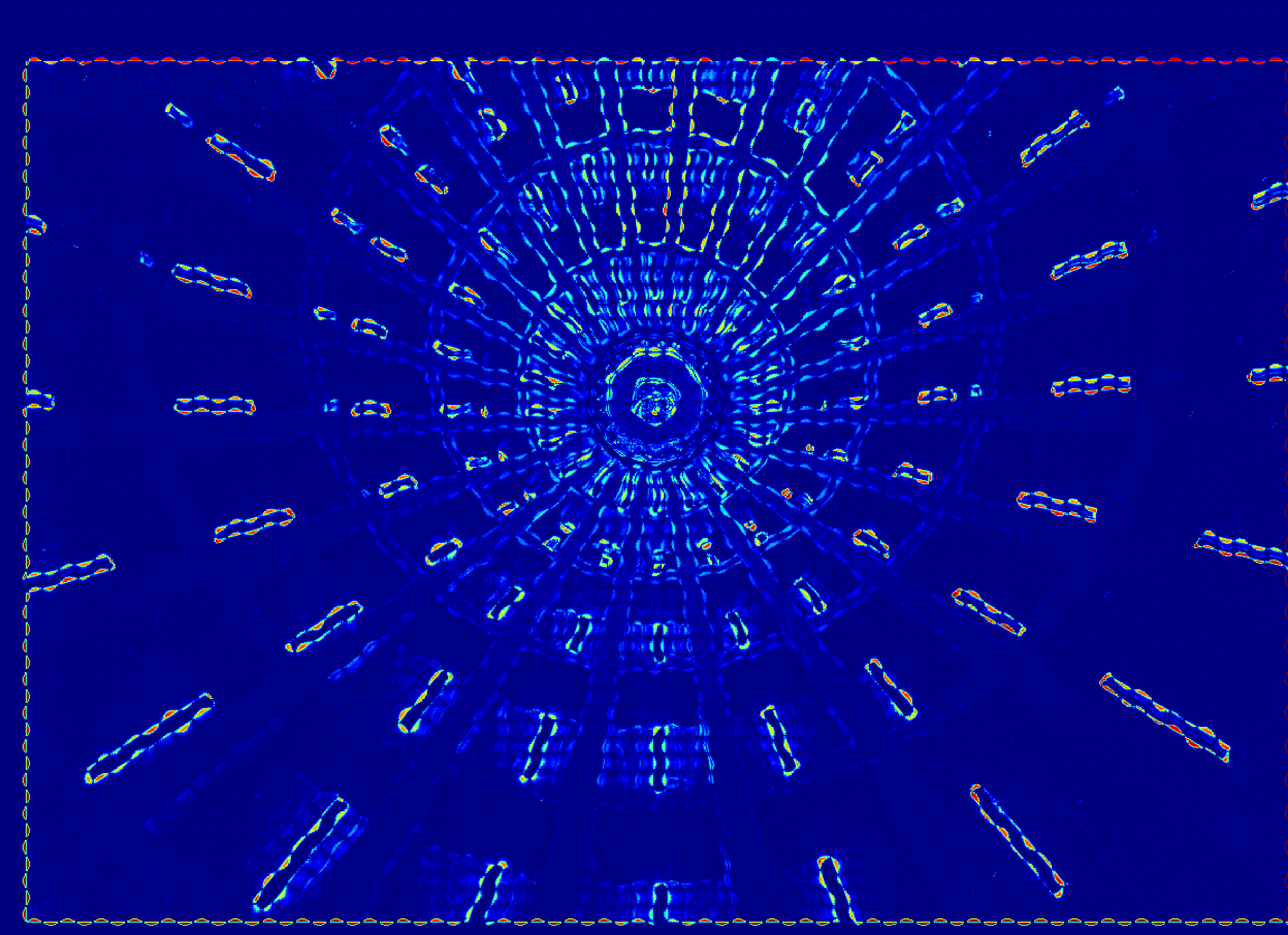} & \includegraphics[width=0.18\linewidth]{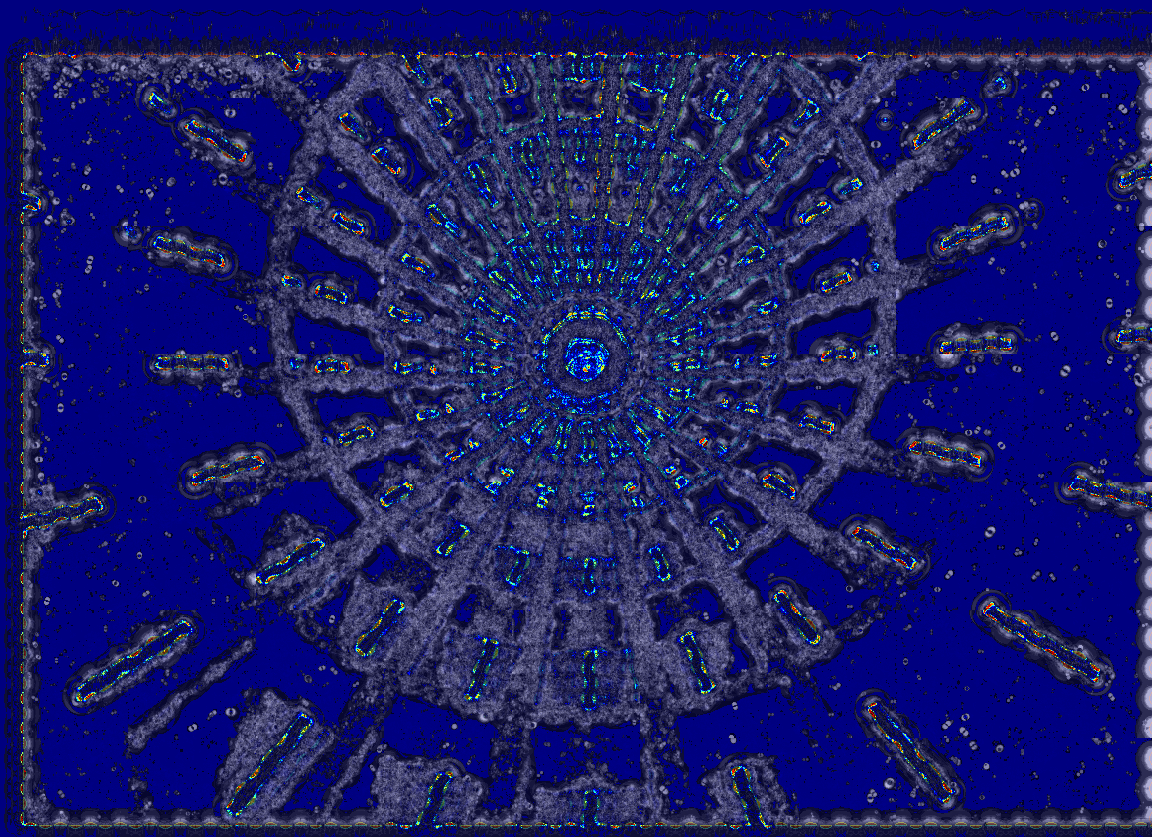} \\
    \includegraphics[width=0.18\linewidth]{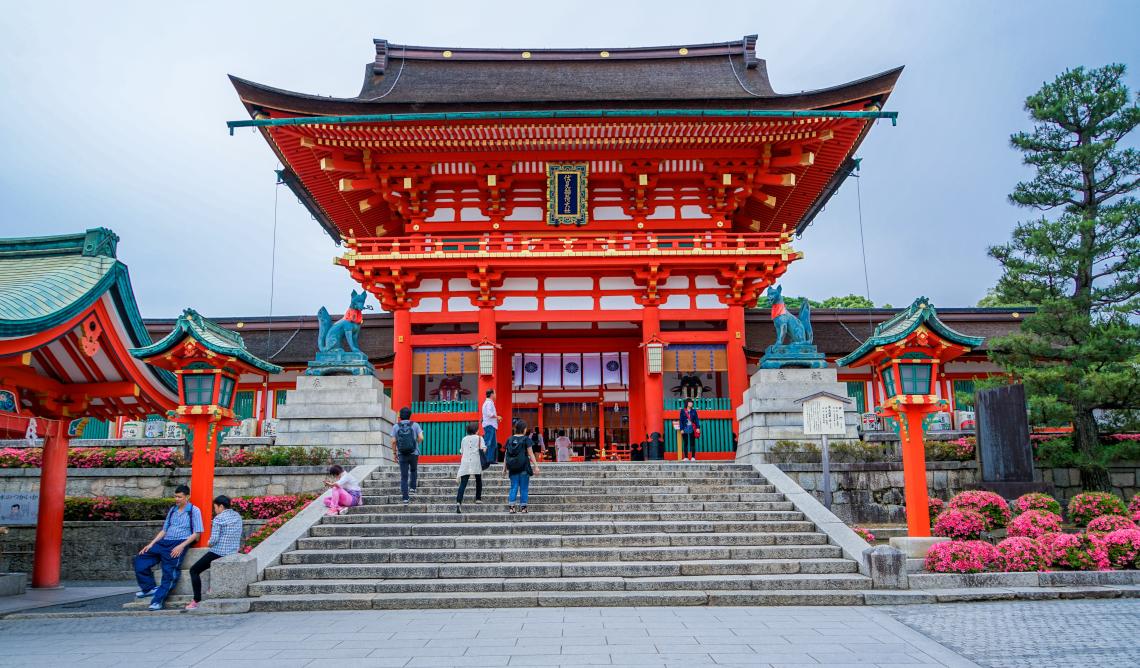} & \includegraphics[width=0.18\linewidth]{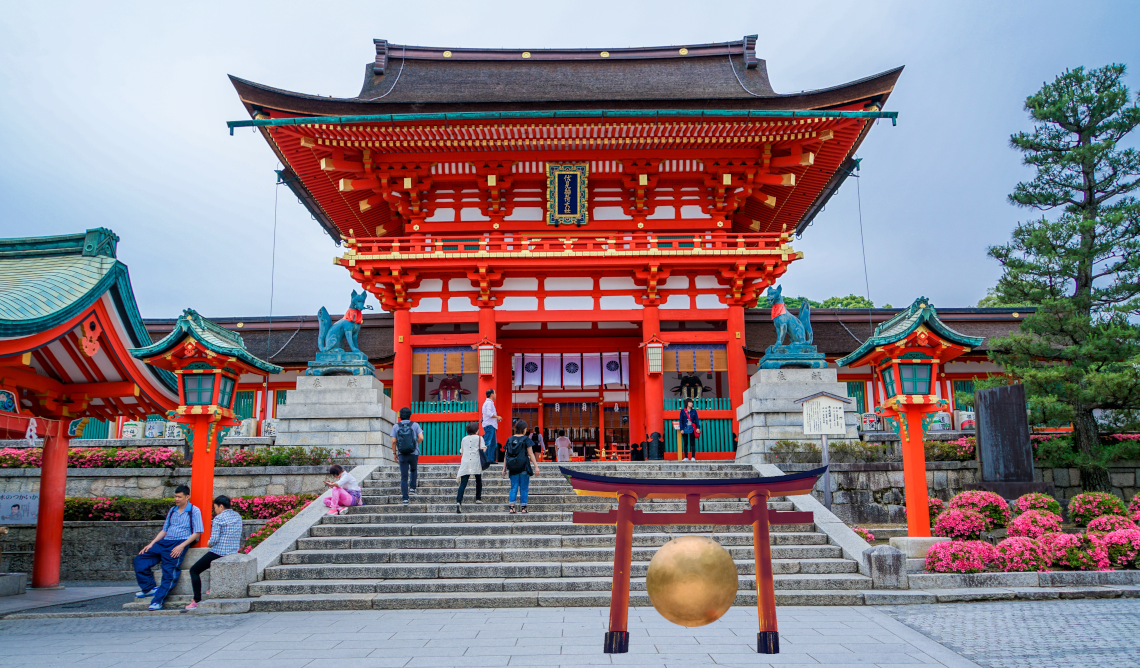}  &     \includegraphics[width=0.18\linewidth]{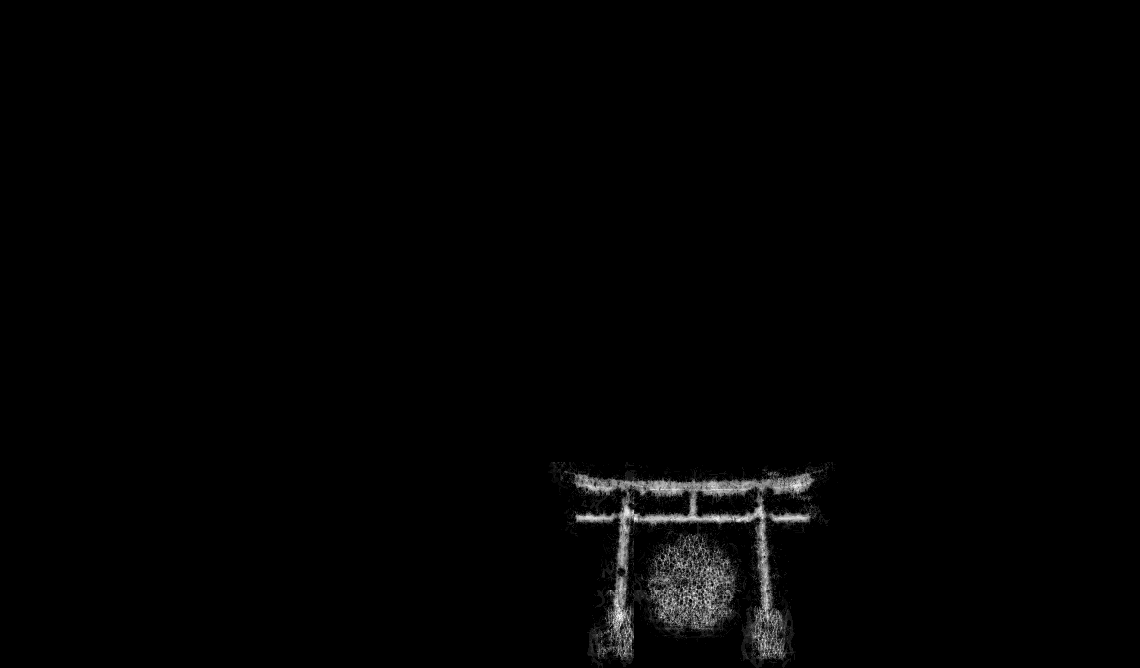} &
    \includegraphics[width=0.18\linewidth]{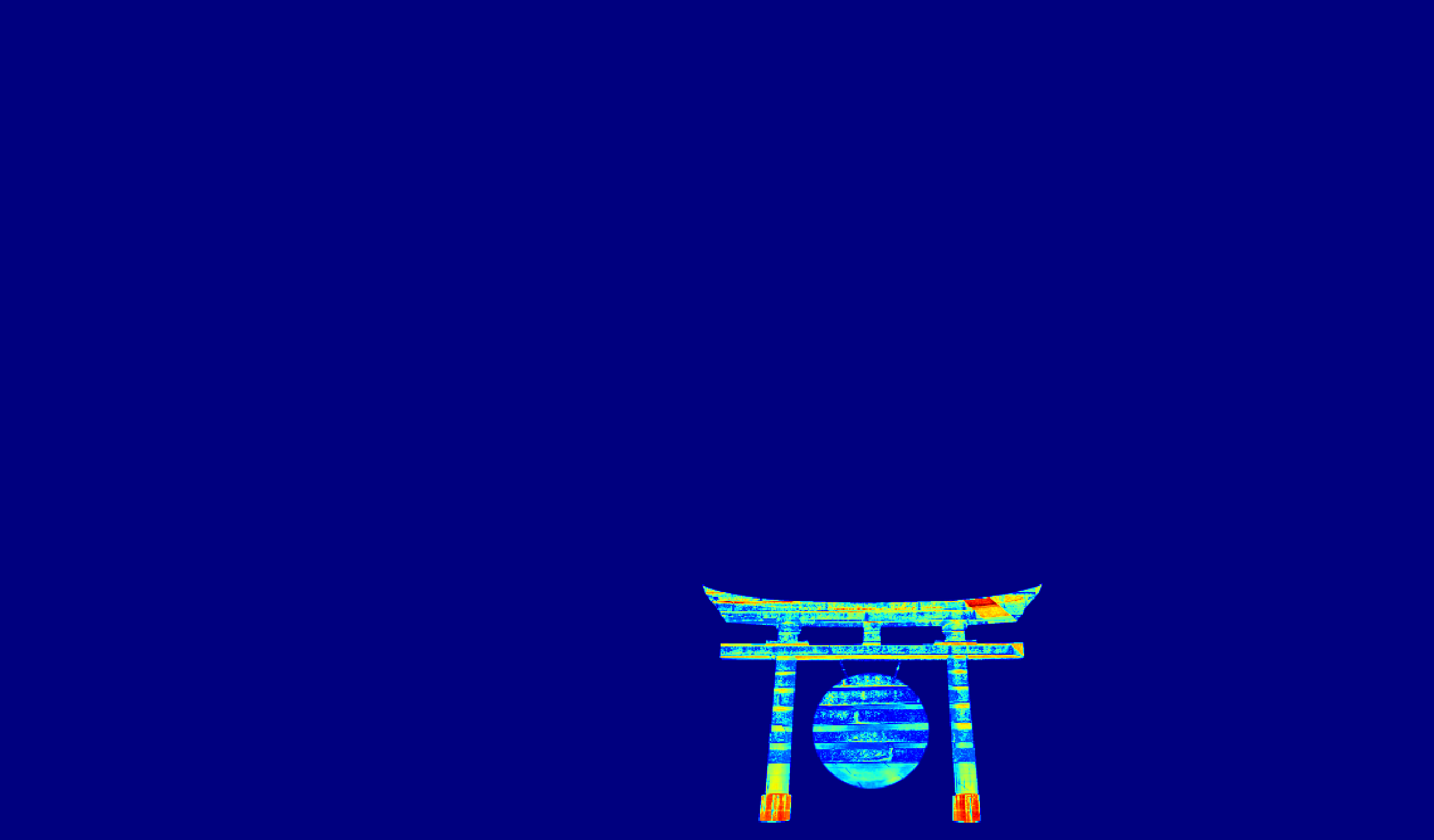} & \includegraphics[width=0.18\linewidth]{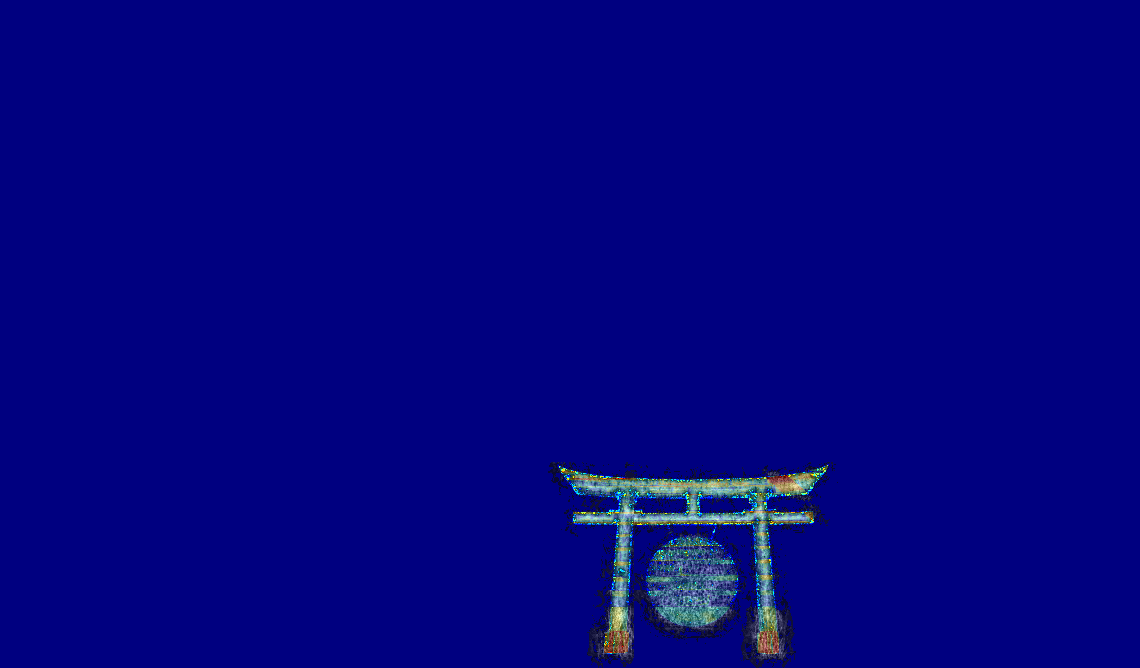} \\
    \includegraphics[width=0.18\linewidth]{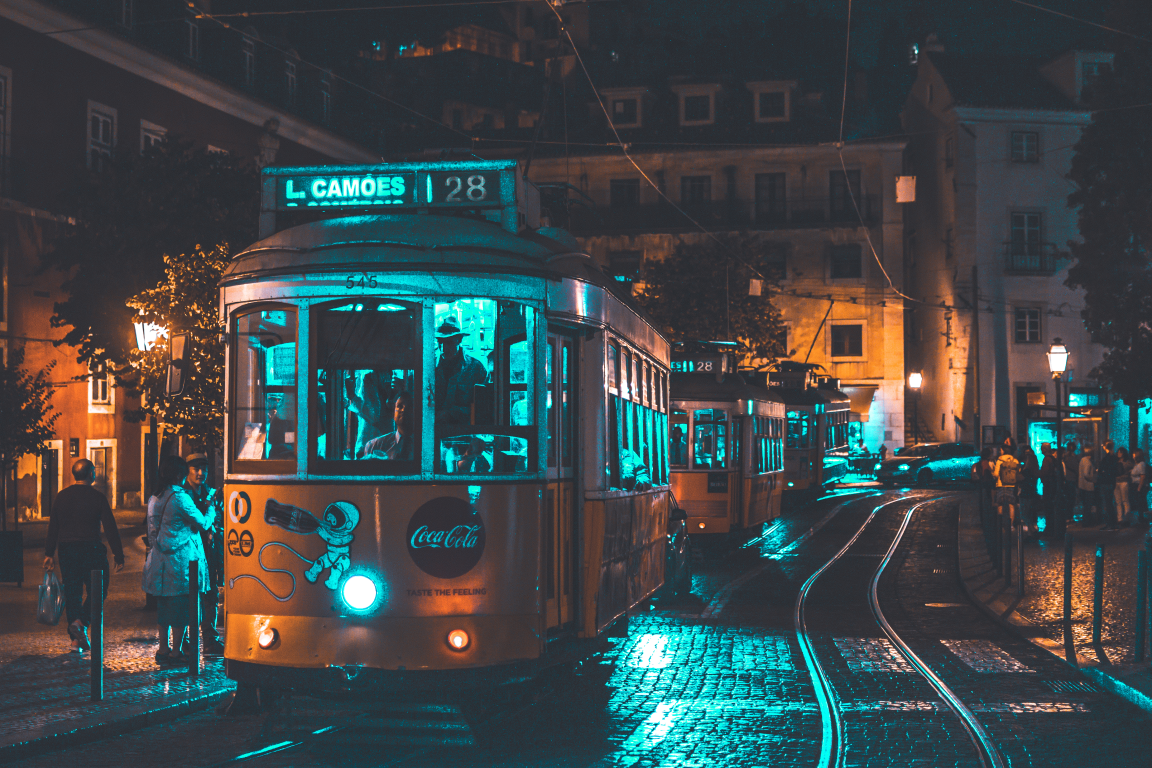} & \includegraphics[width=0.18\linewidth]{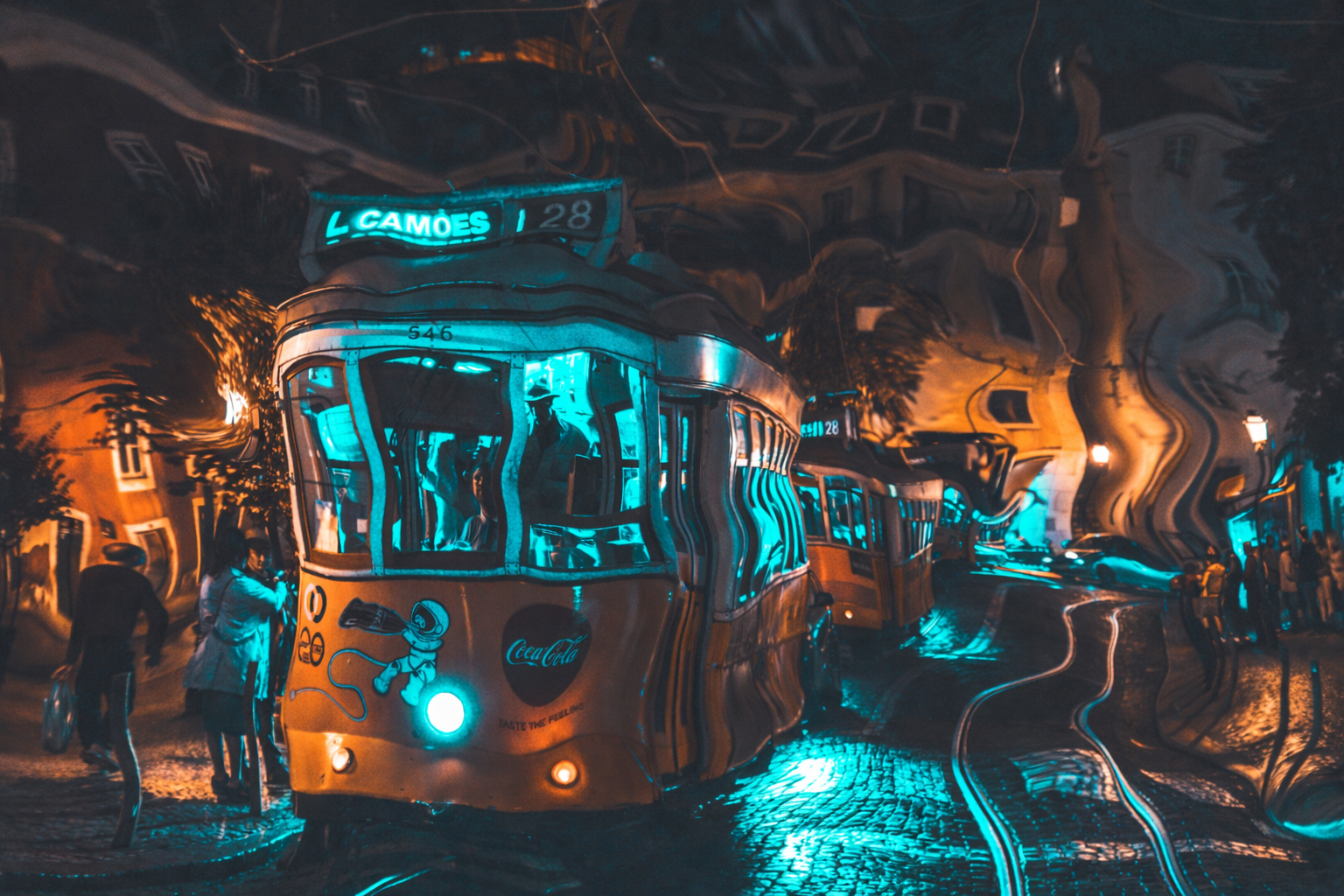}  &     \includegraphics[width=0.18\linewidth]{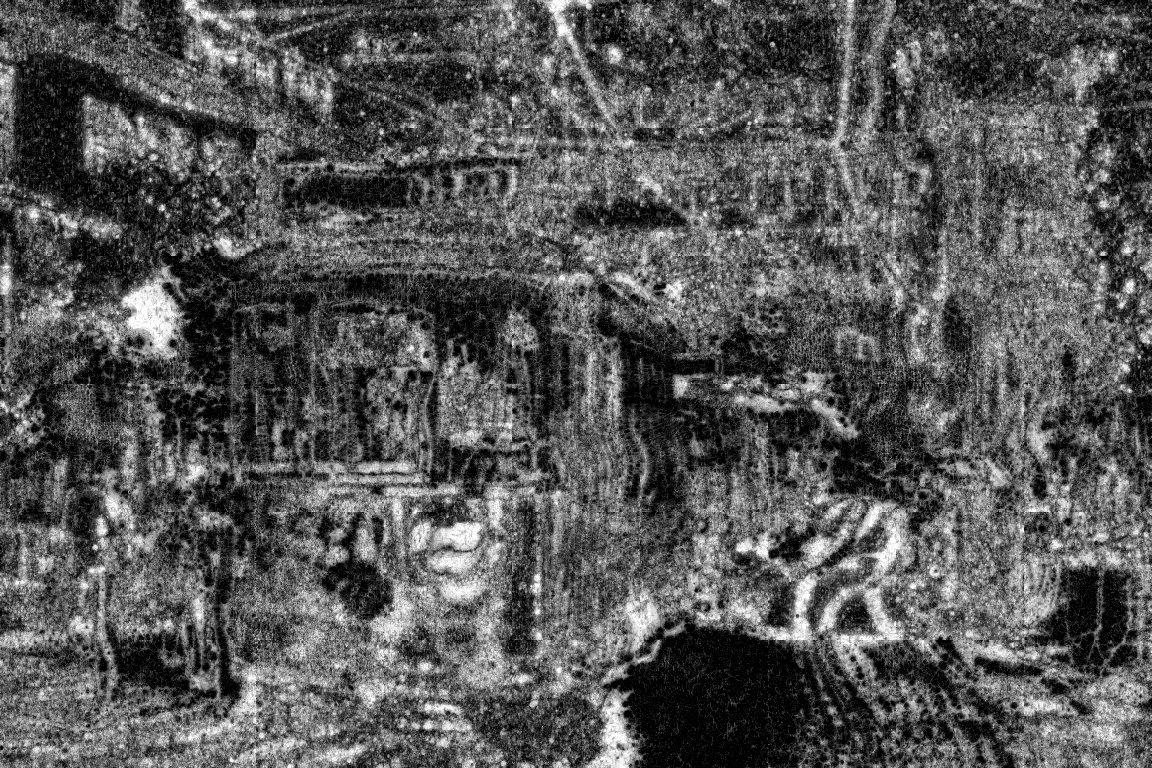} &
    \includegraphics[width=0.18\linewidth]{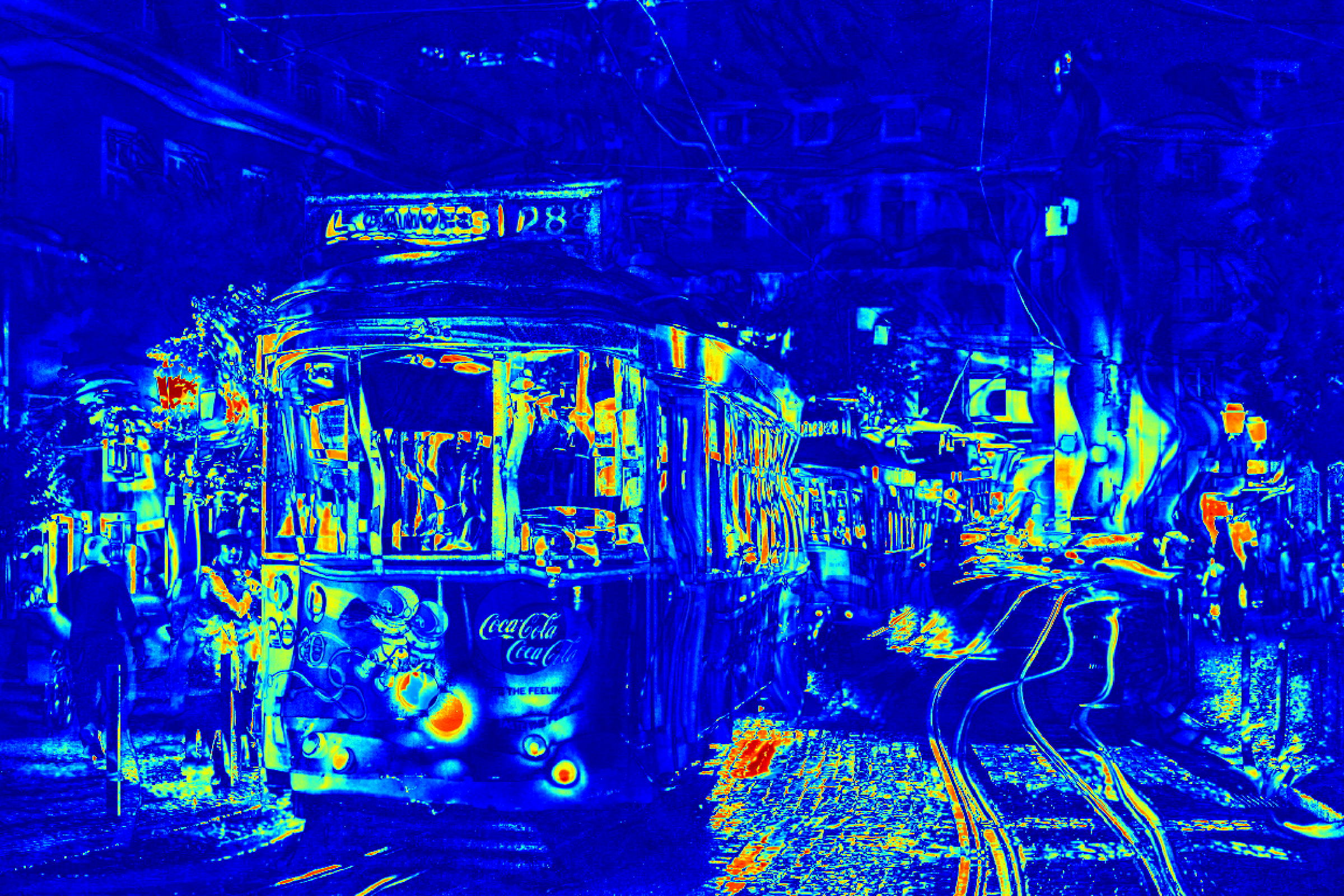} & \includegraphics[width=0.18\linewidth]{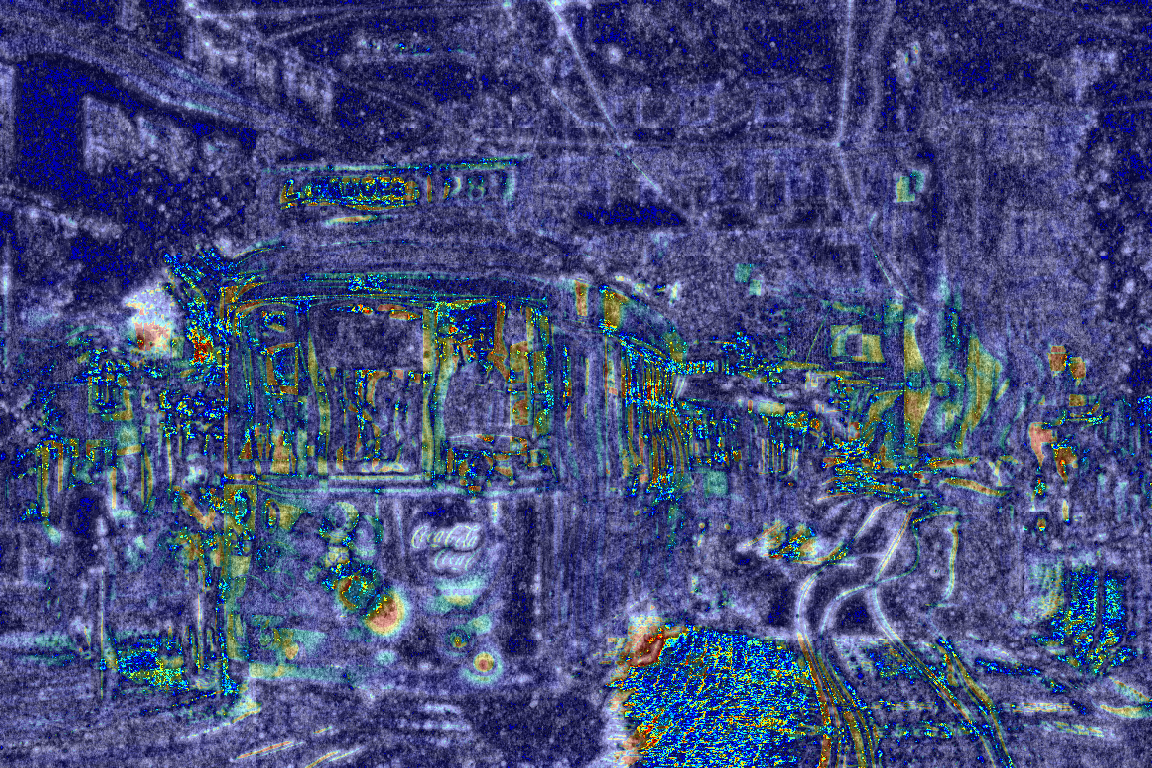}
\end{tabular}
\caption{Examples of explainability provided by EDOKS. The first four rows show images from the LIU4K\_v2 dataset \cite{Liu4K}, and the last row shows a publicly licensed image. The first two columns contain pairs of images to be compared: the original image and its distorted version. The third column shows the map of average magnitude differences extracted for the EMD term. The fourth column shows heatmaps of color differences calculated from the OK term. The fifth column shows the combination of the two maps provided by the two terms. EDOKS scores for these five images: 4.58; 2; 4.18; 26.65 and 3.99 }
\label{fig:explaination}
\end{figure*}

The SOTA metrics discussed in this paper provide a similarity score, but they were not designed to justify the provided score. 
In fact, studies show that metrics such as PSNR do not consistently readjust their score when the degree of distortion is altered \cite{ebrahimi_jpeg_2004}. Without a heatmap highlighting the perceptual differences between two images according to the IQA metric, it is difficult for users to understand what affected the similarity score. This forces users to conduct appropriate experiments to extrapolate the discriminatory and perceptual behavior of the IQA metric.

Meanwhile, we developed EDOKS with the aim of providing a perceptual score and maps of interest showing the discriminative areas that the metric extracts and uses to determine its score. 

Figure \ref{fig:explaination} shows examples of EDOKS's ability to identify discriminatory areas with the greatest difference between two images $X$ and $Y$, providing maps of interest based on differences in shapes and colors.

The EMD difference column highlights the points where the two images differ in shape. The difference in magnitude $|F|$ between the two images is shown, averaging the magnitude differences across all scales $s$ and orientations $o$.

Meanwhile, the OK heatmap column highlights the color difference $\Delta E(\hat{X}_{i,j}, \hat{Y}_{i,j})$ between all pixels of the two images in the Oklab space.

The last column shows an overlay of the two maps calculated on shapes and colors, providing a general overview of the elements that influenced the EDOKS calculation.

The areas that influenced the score coincide with the distorted areas between the two images. Furthermore, the analysis shows that EDOKS is not influenced by distractors because it never highlights the subject of the image and does not present any type of distortion. In fact, the EDOKS map in the first row shows that EDOKS does not identify any color or texture alterations on the butterfly.

\begin{figure*}[tb!]
\centering
\begin{tabular}[c]{ccccc}
 $X$ & $Y$ & $\Delta |F|_{XY}$ & $\Delta E_{XY}$ & $\Delta O$ \\
    \includegraphics[width=0.18\linewidth]{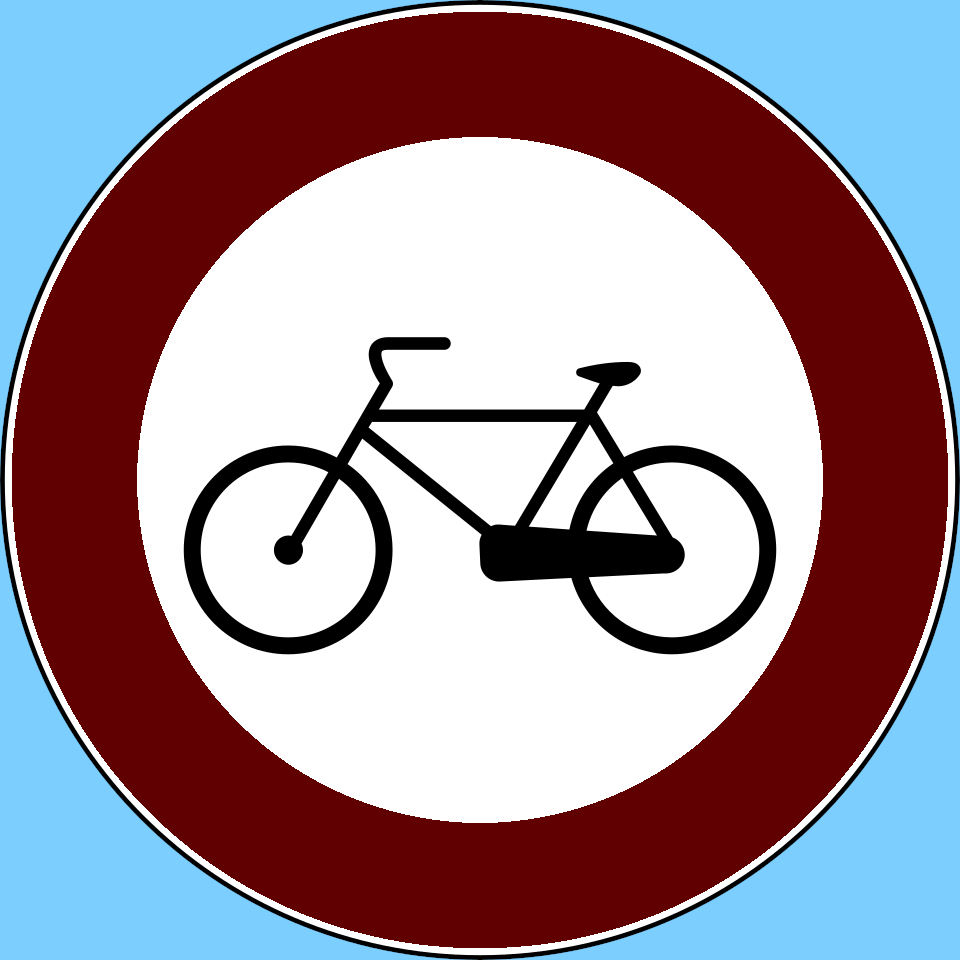} & \includegraphics[width=0.18\linewidth]{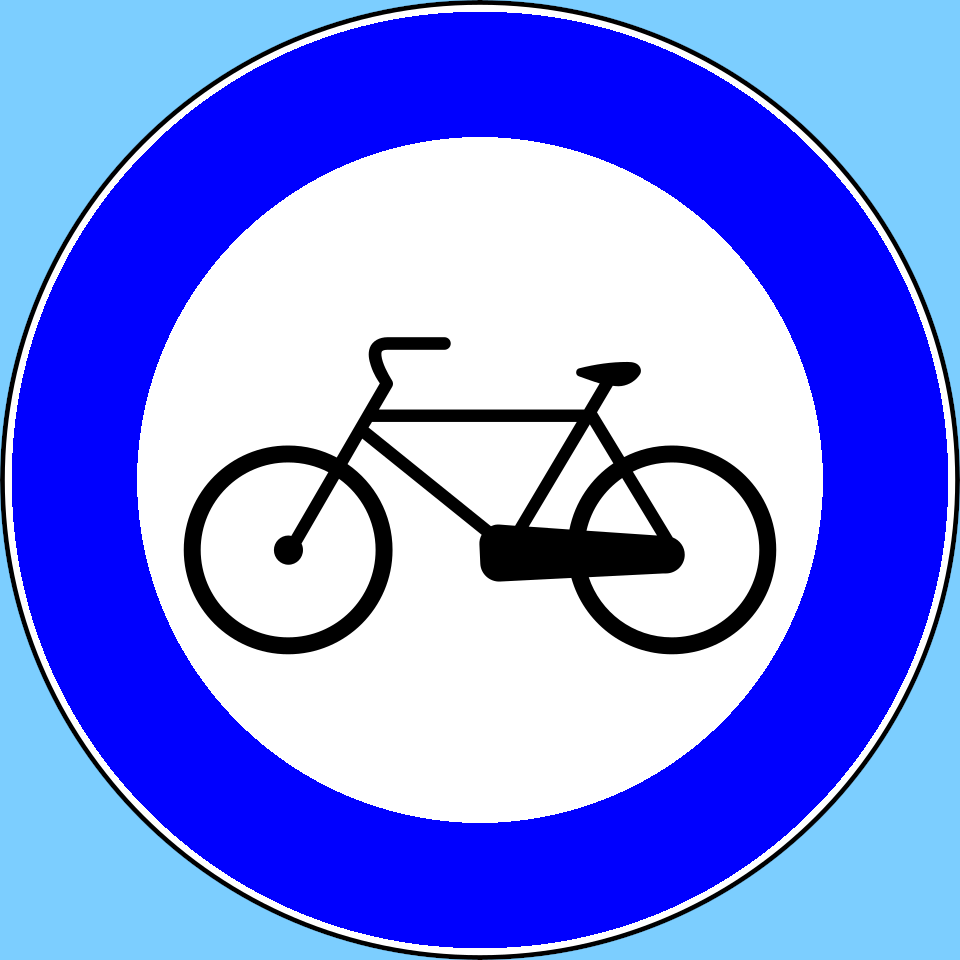}  &     \includegraphics[width=0.18\linewidth]{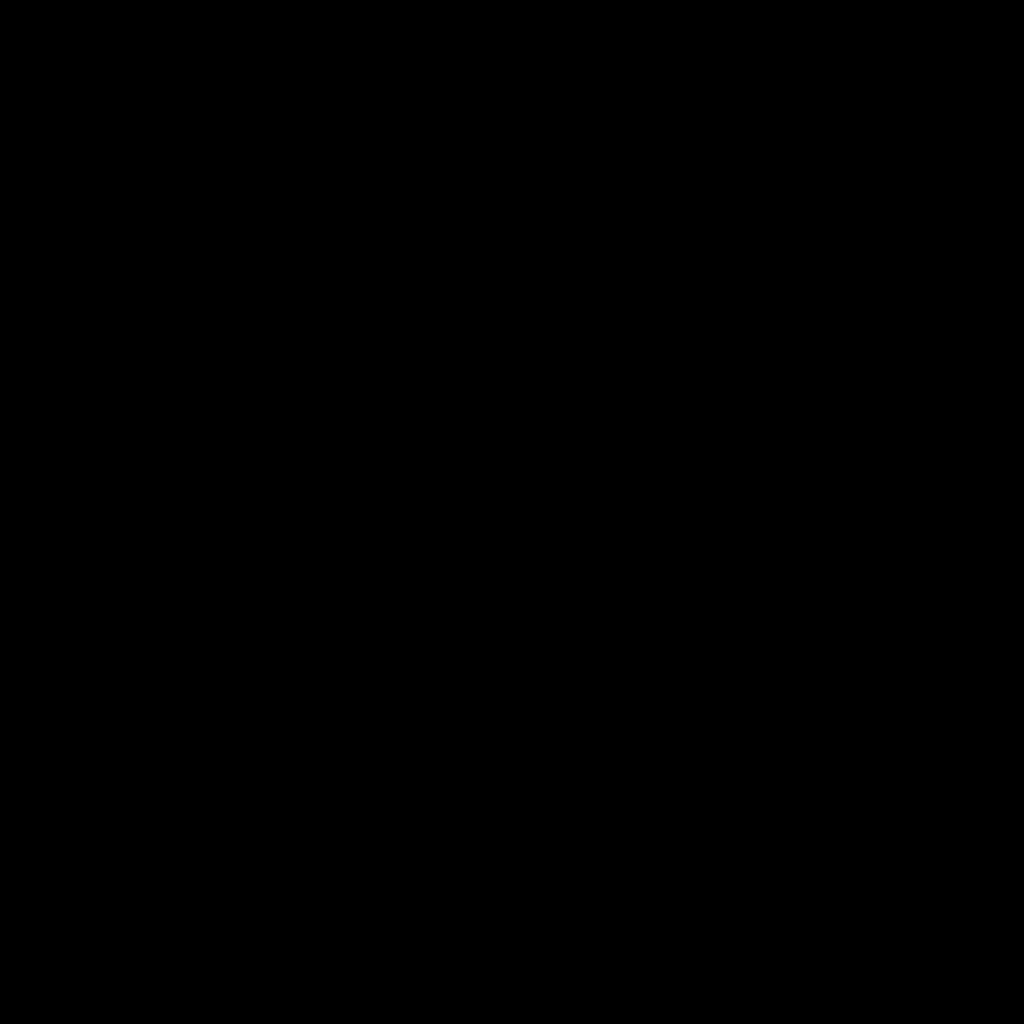} &
    \includegraphics[width=0.18\linewidth]{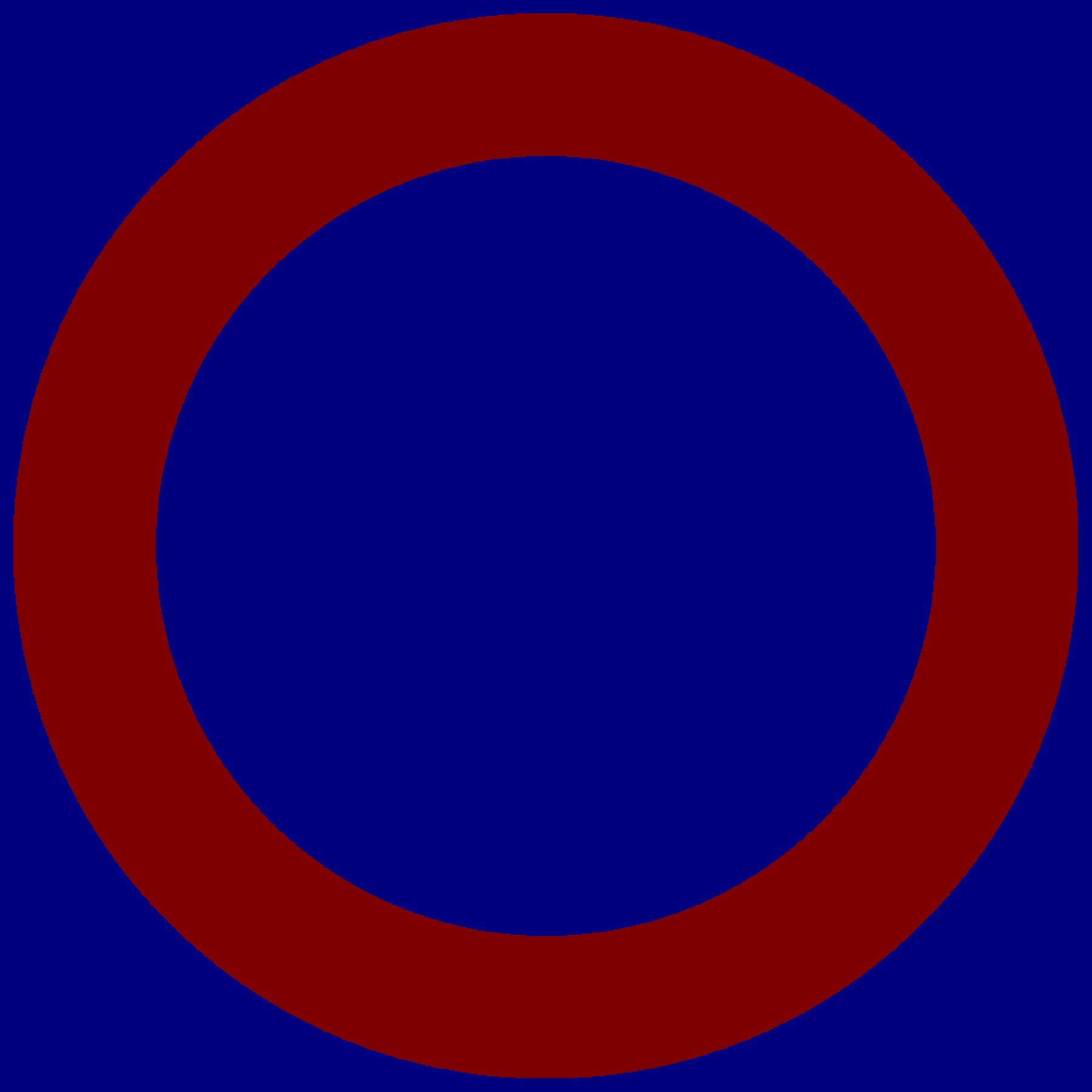} & \includegraphics[width=0.18\linewidth]{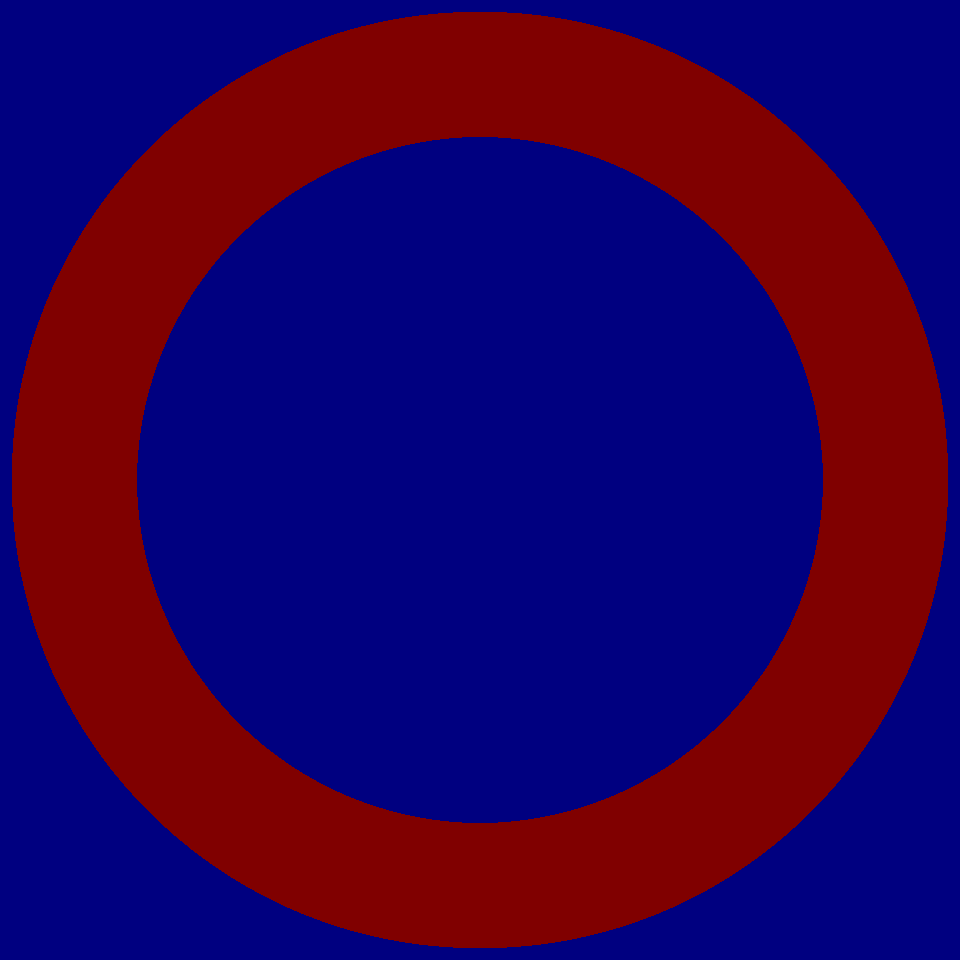}
\end{tabular}

\vspace{2mm}
\footnotesize
\resizebox{\textwidth}{!}{
\begin{tabular}{ccccccccccccccccccc}
EDOKS &       
PSNR &        
SSIM &        
MDSI &        
DSS &         
VSI &         
MS\-GMSD &    
SR\-SIM &     
FSIM &        
GMSD &        
HaarPSI &     
IW\_SSIM &    
MS\_SSIM &    
VIF &         
PSNR\-HVSM &  
LPIPS &       
PIEAPP &      
DISTS &       
TOPIQ \\
  13.511 &
  75.569 &
  1.000 &
  0.504 &
  1.000 &
  0.987 &
  0.000 &
  1.000 &
  1.000 &
  0.000 &
  0.916 &
  1.000 &
  1.000 &
  1.000 &
  100.000 &
  0.195 &
  2.881 &
  0.113 &
  0.597
\end{tabular}}
\caption{Examples of explainability provided by EDOKS in a borderline case. The heatmaps provided by EDOKS and the scores calculated by other metrics are shown.}
\label{fig:bike_signal}
\end{figure*}

Another interesting point is the necessity of using both EMD and OK terms in the EDOKS equation. Alterations were created ad hoc in some areas of the image to stress and challenge the metric. Some areas are more noticeable with one term than the other.

For instance, the difference between the green and magenta backgrounds in the first row is more noticeable with the term "OK" than with the term "EMD", which, like some metrics that convert color to grayscale intensity by assuming similar values, loses information about the clear difference between the two colors. Similarly, the OK term fails to capture geometric differences while the EMD term does. 

This demonstrates that using a single term is necessary to capture a type of distortion but insufficient for a comprehensive analysis of perceptual similarity. 

The second row presents Gaussian smoothing and image negation as distortion. From the maps provided by EDOKS, it can be seen that the metric detects many differences in shape. This is caused by Gaussian smoothing, which erodes all the edges in the image, while the color differences are due to image negativization. It can be seen that all the dark areas become super bright, becoming the areas where the metric highlights the greatest differences.

In the third row, there is a sinusoidal wave warping distortion. From the maps, it can be seen that the texture map highlights the distortion on the shapes, while the color heatmap shows that the color difference highlighted by the metric is only present on the distorted edges and that the image does not show any other color distortions.

The fourth row contains an occlusive element that the metric identifies in both heatmaps. This highlights the only difference between the two images.

For the fifth row, we tested EDOKS with a nocturnal image featuring elastic warping. The maps of interest demonstrate that EDOKS can highlight differences and justify the provided score in terms of both shapes and colors.

In Figure \ref{fig:bike_signal}, the EDOKS metric and the other metrics discussed in this document were tested on a freely licensed image for road signs.

Image $X$ is a no-entry sign for bicycles because it contains a circle with a red border, while image $Y$, although not completely blue, resembles a mandatory sign used for cycle paths due to the presence of this color. The opposite meanings of the two road signs would cause a serious road safety problem.

We show the scores of all the metrics, as well as the EDOKS maps of interest. Our goal is to demonstrate the usefulness of having visible evidence of what the metric considers when providing a score. 

Many metrics, such as SSIM, PSNR-HVSM, VIF, DSS, and SSIM variants, provided a maximum similarity score. This means that, for these metrics, the two images are identical, even if they have a color distortion that is obvious to the human eye. So why do they perform this way?

Without a module that makes them transparent and shows their behavior, it is difficult to determine. Scrutinizing their operations, they convert color images to grayscale using the ITU-R BT.601 standard, so that only the luminance is used. The problem is that many color combinations, such as this shade of red and blue, are converted to the same gray intensity, resulting in two identical images for metrics that adopt this conversion.

Meanwhile, EDOKS provides a similarity score of 13.511, and $\Delta |F|_{XY}$ shows that it has not detected any distortions in the shapes, while the $\Delta E_{XY}$ heatmap highlights color differences in the area where the distortion is applied. We believe that this helps users to evaluate their images beyond simply using numerical scores.

\section{Conclusions}

In this paper, we aimed to highlight the difficulties of SOTA FR-IQA metrics in the perceptual evaluation of images. Therefore, we proposed a new similarity metric, EDOKS, which solves these problems and stands as an alternative to existing metrics in the literature. 

Unlike other similarity indices based on the use of black-box models, such as the LPIPS metric, our proposed metric is interpretable. This means that it is possible to understand why EDOKS provides a similarity score by analyzing the individual terms EMD and OK, which are themselves interpretable because they do not use black-box models, but provide information, respectively, on the dissimilarity of the textures and colors of the compared images through filtering algorithms.

For future developments we believe it may be useful, depending on contexts, to modify parameters such as patch size $p$, Gabor filtering, clustering algorithm or distance used in OK. 
We would also investigate the addition of terms to analyze other types of characteristics in the image that could be indispensable in contexts other than the shape distortions we have analyzed in this work. Furthermore, we evaluate the possibility of exploring the Pareto front for $\alpha$ optimization in Eq. \ref{eq:alpha}.

Following the experiments and discussions, we recommend using EDOKS; we have demonstrated that, especially in the presence of geometric distortion, the EDOKS metric is a more valid perceptual metric compared to the others.

\section*{Code Availability}
We hope that our perceptual metric will be of interest to the scientific community. 

To support the reproducibility of our results, the implementation of the proposed similarity metric is made publicly available at: \url{https://github.com/antdimarino/EDOKS}

\section*{Acknowledgments}
This work was supported by the Future Artificial Intelligence Research (FAIR) project (PE0000013 - CUP B53C22003630006), Spoke 3 - Resilient AI, within the National Recovery and Resilience Plan (PNRR) of the Italian Ministry of University and Research (MUR).

\bibliographystyle{IEEEtran}
\bibliography{IEEEabrv, bibliography}

\vspace*{-3\baselineskip}
\begin{IEEEbiography}[{\includegraphics[width=1in,height=1.25in,clip,keepaspectratio]{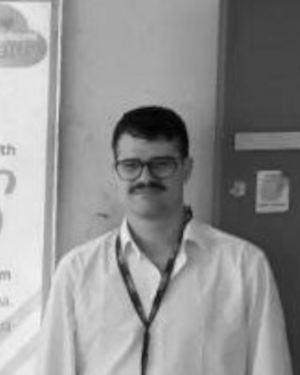}}]{Antonio Di Marino} received the BS degree in Computer Science and the MS degree in Applied Computer Science (Machine Learning \& Big Data) from the University of Naples Parthenope, Naples, Italy, in 2020 and 2023, respectively. He is currently working toward the Italian National PhD program in Artificial Intelligence for Agrifood and Environment with the University of Naples ``Federico II" in collaboration with the Institute for High-Performance Computing and Networking (ICAR) of the National Research Council of Italy (CNR).\end{IEEEbiography}
\vspace*{-2\baselineskip}

\begin{IEEEbiography}[{\includegraphics[width=1in,height=1.25in,clip,keepaspectratio]{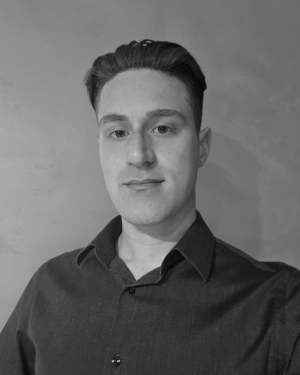}}]{Vincenzo Bevilacqua}
received the B.S. degree in Computer Science and the M.S. degree in Applied Computer Science (Machine Learning \& Big Data) both from the University of Naples Parthenope in 2020 and 2023, respectively. In 2023, he started the Italian National PhD course in Artificial Intelligence.
His research interests include machine learning and deep learning techniques to solve problems of computer vision, such as object detection, image segmentation, tracking video and generative AI.\end{IEEEbiography}
\vspace*{-3\baselineskip}

\begin{IEEEbiography}[{\includegraphics[width=1in,height=1.25in,clip,keepaspectratio]{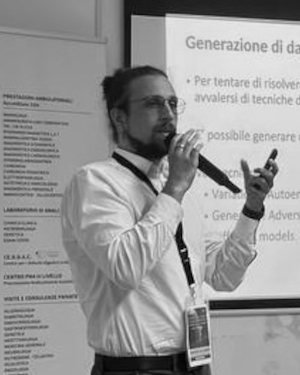}}]{Emanuel Di Nardo} is an Assistant Professor at University of Naples Parthenope and member of the Computational Intelligence \& Smart Systems Lab. His main interests are deep neural networks modeling in Computer Vision. His interests, also, spread on multiple fields, from generative methodologies to building deep neuro-fuzzy architecture to application in interdisciplinary fields like Marine Engineering, Biology and Medical Applications, with a deep focus on explainable AI.\end{IEEEbiography}
\vspace*{-3\baselineskip}

\begin{IEEEbiography}[{\includegraphics[width=1in,height=1.25in,clip,keepaspectratio]{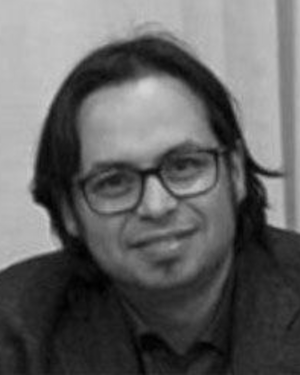}}]{Angelo Ciaramella} is Full Professor at the Department of Science and Technology of the University of Naples “Parthenope”. He is Head of the Computational Intelligence \& Smart Systems Lab (cisslab.uniparthenope.it) and Director of the Apple Foundation Program Parthenope (iosdeveloperacademy.uniparthenope.it). The main research interests are on foundational models of Computational Intelligence, Machine Learning and Data Mining and applications in different fields, as signal processing (i.e., audio, streaming, astrophysical and geological), computer vision and bioinformatics. He is associate editor of international journals (e.g., Information Sciences), he has been co-editor of books and guest editor of Special Issues. He is in the steering committee of WILF conference, has been general Chair (BBCC2023, ITADATA2023, PDP2023, WILF2021, IDCS2019), technical chair (CIBB2018), organizer and chair of Special Sessions (e.g., IJCNN, EAIS, CIBB, WIRN, Fuzz-IEEE, NAFIPS), and he is in the Program Committee (e.g., CIBB, EAIS, Fuzz-IEEE, WIRN, GCIS, ICIC, AI2IA) of international conferences.  He is a Senior Member of IEEE and member of IEEE Computational Intelligence Society, IEEE Signal Processing, SIREN, GIRPR and AIxIA.\end{IEEEbiography}
\vspace*{-3\baselineskip}

\begin{IEEEbiography}[{\includegraphics[width=1in,height=1.25in,clip,keepaspectratio]{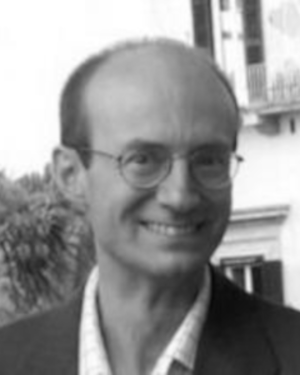}}]{Ivanoe De Falco} is a Research Director at the Institute for High-Performance Computing and Networking (ICAR) of the National Research Council of Italy (CNR). His expertise lies in computational intelligence, machine learning, evolutionary algorithms, swarm intelligence, and parallel and distributed computing. Currently, he is focusing on interpretable and explainable AI techniques within the FAIR (Future Artificial Intelligence Research) project funded by the Italian National Recovery and Resilience Plan (PNRR). Ivanoe De Falco has authored over 200 peer-reviewed publications and has been listed in Stanford University’s Top 2\% Scientists ranking from 2021 to 2024. \end{IEEEbiography}
\vspace*{-3\baselineskip}

\begin{IEEEbiography}[{\includegraphics[width=1in,height=1.25in,clip,keepaspectratio]{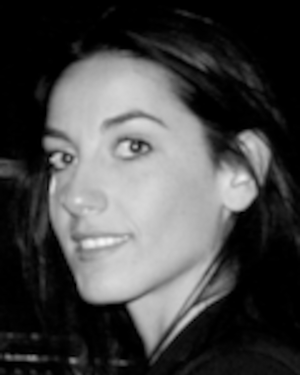}}]{Giovanna Sannino} is a Senior Researcher at the Institute for High Performance Computing and Networks (ICAR) of the National Research Council of Italy (CNR). Her skills and research interests concern signal processing, ML/AI techniques for eHealth applications with an emphasis on interpretable and explainable methods. She is currently a member of several national and international working groups, such as the IEEE 11073 Personal Health Device Working Group and the World Federation on Soft Computing (WFSC). She has been involved in various regional and national projects, also with managerial duties, and has organized several international conferences and workshops. Author of more than 100 scientific peer-reviewed articles, of an international patent, and guest editor of numerous international journals, she is currently Executive Editor for the Biomedical Signal Processing and Control Journal, Associate Editor for IEEE Transactions on Affective Computing and for Applied Soft Computing Journals, and Editor Member of Engineering Applications of Artificial Intelligence Journal.\end{IEEEbiography}
\vspace*{-2\baselineskip}

\end{document}